\pdfoutput=1
\documentclass[10pt,twocolumn,letterpaper]{article}
\makeatletter
\@namedef{ver@everyshi.sty}{}
\makeatother
\usepackage{cvpr}              

\usepackage{graphicx}
\usepackage{amsmath}
\usepackage{amssymb}
\usepackage{booktabs}
\usepackage{algorithm}
\usepackage{algorithmicx}
\usepackage{algpseudocode}
\usepackage{multirow}
\usepackage[table,xcdraw]{xcolor}
\usepackage{booktabs}
\usepackage{verbatim}
\usepackage{color, colortbl}
\usepackage{bm}
\usepackage{enumitem}
\usepackage{pifont}
\usepackage[accsupp]{axessibility}
\usepackage{multirow}
\usepackage{makecell}
\usepackage{adjustbox}
\usepackage{bbding}
\usepackage{tikz}
\usepackage{pgfplots}
\usepackage{ulem}
\usepackage{mathtools}
\usepackage{pgfplotstable}
\usepackage{setspace}

\usepackage[pagebackref,breaklinks,colorlinks]{hyperref}

\usepackage[capitalize]{cleveref}
\crefname{section}{Sec.}{Secs.}
\Crefname{section}{Section}{Sections}
\Crefname{table}{Table}{Tables}
\crefname{table}{Tab.}{Tabs.}

\newcommand{\cmark}{\ding{51}}%
\newcommand{\xmark}{\ding{55}}%

\definecolor{Gray}{gray}{0.9}
\newcommand{\best}[1]{\color{red}\textbf{#1}}
\newcommand{\second}[1]{\color{blue}\textbf{#1}}

\DeclarePairedDelimiter\abs{\lvert}{\rvert}%
\DeclarePairedDelimiter\norm{\lVert}{\rVert}%
\def\cvprPaperID{9868} 
\def\confName{CVPR}
\def\confYear{2023}

\newcommand\blfootnote[1]{%
  \begingroup
  \renewcommand\thefootnote{}\footnote{#1}%
  \addtocounter{footnote}{-1}%
  \endgroup
}

\pgfplotsset{compat=1.9}
\usetikzlibrary{pgfplots.groupplots}

\begin{document}
\setlength{\abovedisplayskip}{3pt}
\setlength{\belowdisplayskip}{3pt}
\title{NIFF: Alleviating Forgetting in Generalized Few-Shot Object Detection\\ via Neural Instance Feature Forging \vspace{-1em}}

\author{{\normalsize Karim Guirguis}$^{1,2 \dagger}$ \hspace{0.1em} {\normalsize Johannes Meier}$^{1 \dagger}$ \hspace{0.1em}  {\normalsize George Eskandar}$^{3}$ \hspace{0.1em}  {\normalsize Matthias Kayser}$^{1}$ 
\hspace{0.1em}  {\normalsize Bin Yang}$^{3}$ 
\hspace{0.1em}  {\normalsize J\"urgen Beyerer}$^{2,4}$\\
{\normalsize Robert Bosch GmbH}$^1$ \hspace{0.2em} {\normalsize Karlsruhe Institute of Technology}$^2$ \hspace{0.2em} {\normalsize University of Stuttgart}$^3$ \hspace{0.2em} {\normalsize Fraunhofer IOSB\vspace{-1em}}$^4$ \\
}
\maketitle


\begin{abstract}
\vspace{-0.5em}
Privacy and memory are two recurring themes in a broad conversation about the societal impact of AI. These concerns arise from the need for huge amounts of data to train deep neural networks. A promise of Generalized Few-shot Object Detection (G-FSOD), a learning paradigm in AI, is to alleviate the need for collecting abundant training samples of novel classes we wish to detect by leveraging prior knowledge from old classes (i.e., base classes). G-FSOD strives to learn these novel classes while alleviating catastrophic forgetting of the base classes. However, existing approaches assume that the base images are accessible, an assumption that does not hold when sharing and storing data is problematic. In this work, we propose the first data-free knowledge distillation (DFKD) approach for G-FSOD that leverages the statistics of the region of interest (RoI) features from the base model to forge instance-level features without accessing the base images. Our contribution is three-fold: (1) we design a standalone lightweight generator with (2) class-wise heads (3) to generate and replay diverse instance-level base features to the RoI head while finetuning on the novel data. This stands in contrast to standard DFKD approaches in image classification, which invert the entire network to generate base images. Moreover, we make careful design choices in the novel finetuning pipeline to regularize the model. We show that our approach can dramatically reduce the base memory requirements, all while setting a new standard for G-FSOD on the challenging MS-COCO and PASCAL-VOC benchmarks.          
\vspace{-2em}
\end{abstract}
\blfootnote{$\dagger$ Authors have equally contributed to this work.}
\blfootnote{Corresponding author: karim.guirguis@de.bosch.com}
 \begin{figure}[t!]
 \centering
 \includegraphics[width=0.87\linewidth]{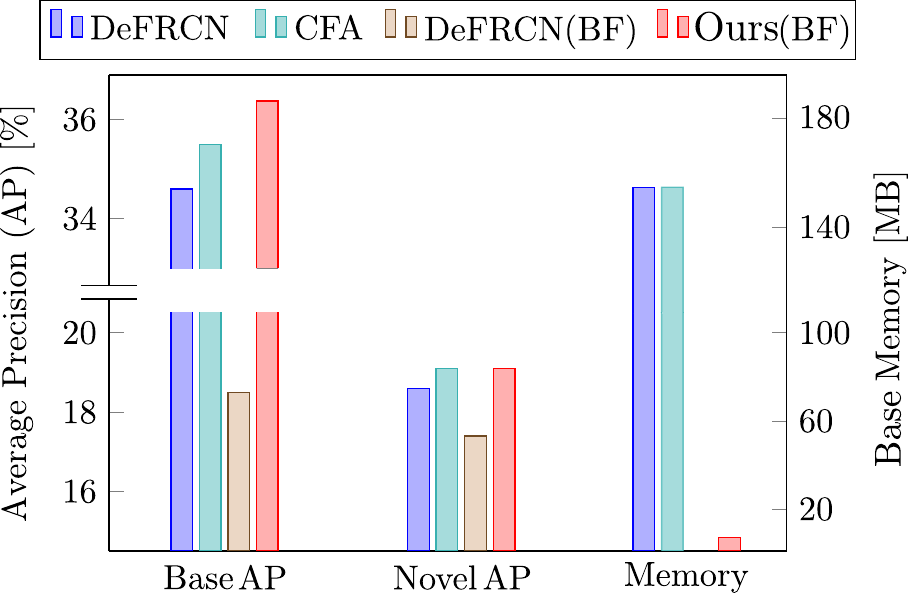}
 \vspace{-0.8em}
\caption{The base memory requirements for G-FSOD are dramatically reduced by our framework, while improving the overall detection performance on MS-COCO ($10$-shot). We only store a lightweight generator that synthesizes deep features for the RoI head. BF denotes base-data free finetuning.\vspace{-2em}} 
 \label{fig:teaser}
\end{figure} 

\section{Introduction}
\label{sec:intro}
Object detection (OD) is an integral element in modern computer vision perception systems (e.g., robotics and self-driving cars). However, object detectors~\cite{R-CNN, FastR-CNN, FasterR-CNN, CascadeR-CNN, SSD, YOLOv1, YOLOv2, RetinaNet} require abundant annotated data to train, which is labor and time intensive. In some applications requiring rare class detection, collecting much data is challenging. Striving to learn in limited data scenarios, few-shot object detection (FSOD)~\cite{fsod_survey} is an uprising field. It mimics the human cognitive ability by leveraging prior knowledge from previous experiences with abundant base data, to rapidly learn novel classes from limited samples. Despite the success of meta-learning~\cite{MetaRCNN, FSOD-RPN, FsDetView, CME, DANA, MetaDetR} and transfer learning~\cite{TFA,MPSR,FSCE,defrcn} paradigms in FSOD, most methods prioritize the detection performance of the novel classes while ignoring the catastrophic forgetting of the base ones. This might lead to critical failure cases in real-life operational perception systems.

 \begin{figure}[t!]
 \centering
 \includegraphics[width=0.852\linewidth]{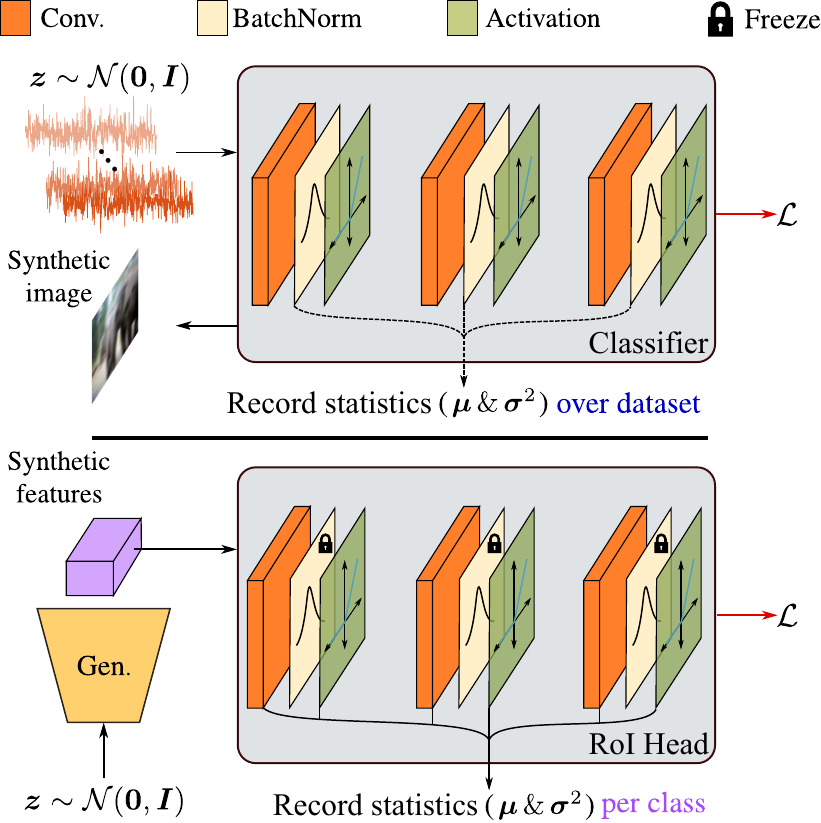}
\caption{\textbf{Top:} Standard DFKD approach via inverting the entire model~\cite{DeepInversion,ABD}. \textbf{Bottom:} An  overview of our proposed approach. We abstractly show a few layers in both models. The main differences are: (1) we synthesize features instead of images, (2) we use a separate generator instead of inverting the model, and (3) we record the class-wise statistics (instead of the full data statistics) before and after the normalization layers in the RoI head.} \vspace{-1.5em}
 \label{fig:intro_contrib}
\end{figure}

To address the aforementioned concern, generalized few-shot object detection (G-FSOD) has been introduced to jointly detect the base and novel classes. One of the first approaches to address the G-FSOD task was TFA~\cite{TFA}, which finetunes the detector using a balanced set of base and novel class samples while freezing the backbone and the region proposal network (RPN). While this has reduced forgetting, the performance on novel classes has dropped significantly. Since then, a plethora of works have attempted to improve the overall detection performance. DeFRCN~\cite{defrcn} proposed a gradient manipulation approach to modify the RPN and RoI head gradients. Retentive R-CNN ~\cite{retentive}, a knowledge distillation approach and CFA~\cite{cfa}, a gradient manipulation approach were proposed to explicitly tackle the catastrophic forgetting of base classes. However, all the above approaches rely on the assumption that base data is available while learning the new classes. This made us raise the following question: \textit{How to alleviate forgetting without base data in case of a memory constraint or privacy concerns restricting the storage and replay of old base data?}

Data-free knowledge distillation (DFKD) is a line of work sharing a similar interest in transferring knowledge without storing raw data. DeepDream~\cite{DeepDream}  pioneered model inversion (MI) work which inverted a pre-trained classifier to generate synthetic images for knowledge distillation. Since then, various works~\cite{DeepInversion, ABD, DIODE} have followed attempting to generate higher-fidelity images and even in a class-incremental setting~\cite{ABD}. Despite its success in image classification, applying DFKD in G-FSOD comes with several challenges. First, most works revolve around generating synthetic images via MI. In the context of OD and G-FSOD, this entails generating images with bounding boxes which inflicts higher computational and memory overhead. Although a recent approach DIODE~\cite{DIODE} has applied MI in OD, it cannot be extended to G-FSOD for the following reason. Similar to all the previously mentioned works in DFKD, DIODE needs the statistics of the BatchNorm(BN)~\cite{batchnorm} layers which are trained on the detection datasets. However, the backbone in G-FSOD is pre-trained on ImageNet and frozen (except the last ResBlock) during the entire training (unfreezing would change the mature parameters and will  reduce the overall performance). Hence, the running means and variances in the BN do not represent the true base data distribution. 


\textbf{Contribution:} In this work, we propose Neural Instance Feature Forging (NIFF),  the first data-free knowledge distillation method for G-FSOD. We aim to alleviate forgetting without storing base data to respect privacy restrictions and reduce the overall memory footprint, as shown in~\cref{fig:teaser}. Our two key insights are as follows. First, we show that the statistics of instance-level RoI head features sufficiently represent the distribution of base classes. Second, we show that a standalone lightweight generator can be trained in a distillation fashion to match the gathered statistics and synthesize class-wise base features to train a G-FSOD model. This stands in contrast to MI approaches which optimize the pre-trained model to synthesize high-fidelity images. Our contributions are summarized as follows:

\begin{enumerate}[topsep=0pt,itemsep=-1ex,partopsep=1ex,parsep=1ex]
    \item We forge instance-level features instead of synthesizing images (\cref{fig:intro_contrib}) as the feature space ($1024 \times 7 \times 7$) is much smaller than the image space ($3 \times 600 \times 1000$). 
    \item Rather than inverting the whole model, instead we design a standalone lightweight generator in the feature space to forge instance-level base features. The generator is able to better capture the feature distribution than the complex MI.
    \item We equip the generator with class-aware heads rather than a class-agnostic one, and we train it to synthesize features with a distribution that matches the class-wise statistics of the pre-trained base RoI head, hence promoting feature diversity. We demonstrate that class-aware heads trained on class-aware statistics outperform a shared model trained on class-wise statistics.  
    \item We dramatically reduce the overall memory footprint by two orders of magnitude while setting a new standard for the overall detection performance in G-FSOD on MS-COCO~\cite{coco} and PASCAL-VOC~\cite{pascalvoc}. For instance, the base images and annotations for MS-COCO dataset ($10$-shot) are $\sim150$MB large, while our generator parameters only occupy $\sim4$MB.\vspace{-0.5em}
\end{enumerate}
\section{Related Works}
\label{sec:related_works}
\textbf{Few-Shot Object Detection.} FSOD strategies can be divided into meta-learning and transfer learning approaches. The meta-learning methods gain knowledge by solving a variety of unrelated tasks~\cite{MetaRCNN, FSOD-RPN, FsDetView, CME, DANA, MetaDetR}. On the other hand, the transfer learning approaches directly finetune on the novel data. On a balanced training set of both base and novel classes, TFA~\cite{TFA} finetunes just the box predictor. By creating multi-scale positive samples as object pyramids, MPSR~\cite{MPSR} addresses the high scale variances by improving the prediction at many scales. DeFRCN~\cite{defrcn} highlights the contradictory goals of the RPN and class-aware ROI head. Accordingly, they remove the RPN gradients and downscale the ROI head gradients flowing to the backbone. However, the main goal of FSOD approaches is to boost the performance on the novel classes.   

\textbf{Generalized Few-Shot Object Detection.} A growing area of study within FSOD is G-FSOD, which focuses on detecting both the base and novel classes. TFA~\cite{TFA} attempts to prevent forgetting by optimizing a balanced set of base and novel classes, whereas ONCE~\cite{once} tackles the issue in an incremental class learning setting using a meta-learning strategy. A transfer-learning-based strategy, Retentive R-CNN~\cite{retentive}, leverages the base-trained model in a distillation-like fashion to alleviate forgetting. To reduce the extra computational and memory costs, CFA~\cite{cfa} developed a new gradient update mechanism to alleviate forgetting based on the angle between gradients for base and novel samples. All of the approaches outlined above, however, are heavily reliant on the availability of annotated base images during the novel training phase. 

\textbf{Data-Free Knowledge Distillation.} DeepDream~\cite{DeepDream} was the first among this line of work to show that discriminative neural networks harbor information that enables the generation of images. It synthesizes an image to provide a high output response for specific classes at different model layers. Later, DeepInversion~\cite{DeepInversion} improved the dreamed image quality by penalizing the distance between statistics of the features, assuming a Gaussian distribution. AlwaysBeDreaming (ABD)~\cite{ABD} minimizes the feature drift over the previous task while finetuning the last classification layer with a cross entropy loss. Recently, Chawla et al.~\cite{DIODE} proposed a non class incremental data-free knowledge distillation approach for YOLOv3 based on MI along with a bounding box sampling scheme. If extended to G-FSOD, the backbone and BN layers would be unfrozen, which would limit the model's capacity to rapidly learn new classes. Moreover, BN layers do not capture the true class-wise distribution.
\vspace{-2mm}

\section{Methodology}
\label{sec:methodology}
We aim to design a G-FSOD pipeline that learns novel classes from scant data while preserving privacy and memory constraints. In this section, we start by formally defining the G-FSOD problem. Then, we revisit the data-free knowledge distillation via noise optimization. Finally, we present our approach, NIFF, which consists of two stages: (1) Feature generator training and (2) Novel training in a distillation fashion via the trained generator.  \vspace{-1mm}
 \begin{figure*}[t!]
 \centering
 \includegraphics[width=0.8\linewidth]{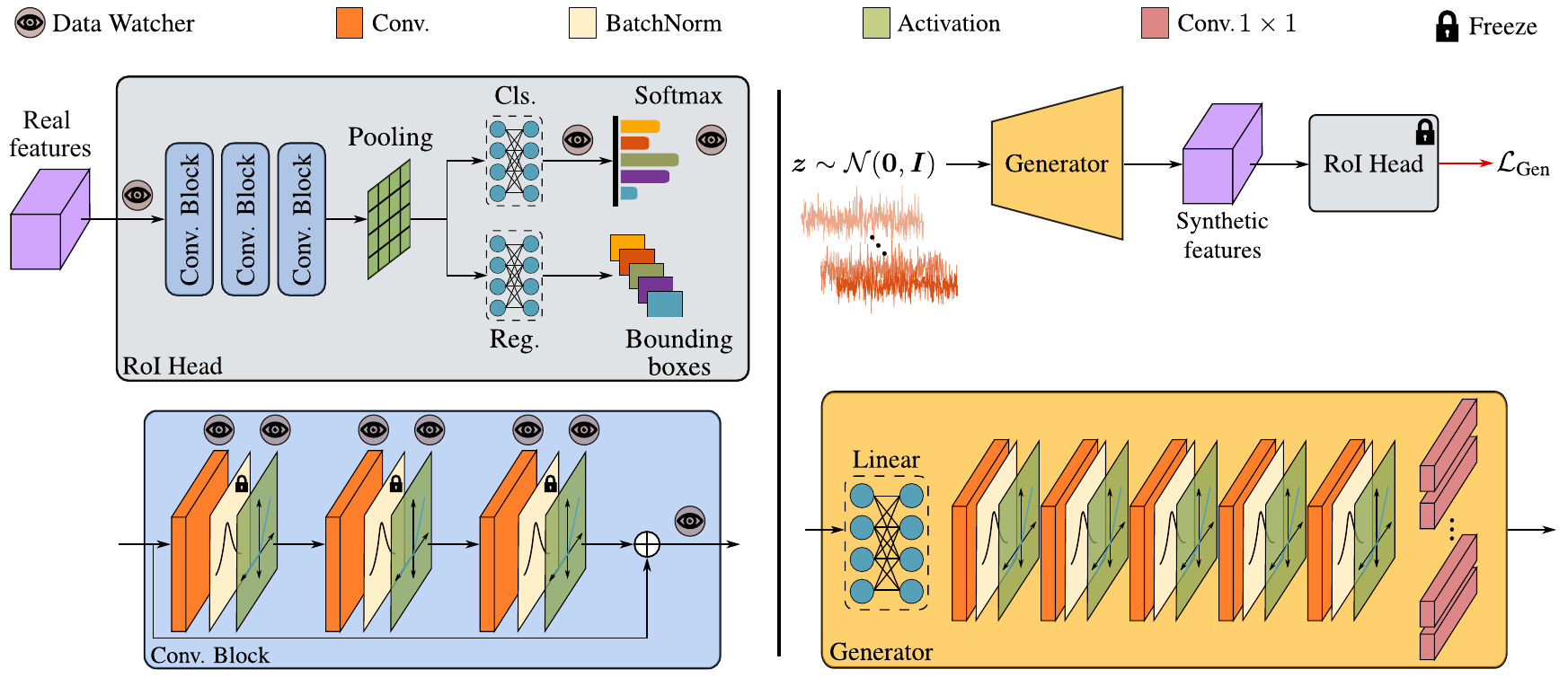}
 \vspace{-0.5em}
\caption{\textbf{Stage I: Feature Generator Training.} \textbf{Left:} A detailed overview of the RoI head to highlight where the features' statistics are gathered using the data watchers. \textbf{Right:} Illustration of the proposed generator training pipeline and the architectural details.} %
\vspace{-1.5em}
 \label{fig:stats}
\end{figure*}


\subsection{Problem Formulation}\vspace{-1mm}
Analogous to FSOD, G-FSOD divides the dataset $\mathcal{D}$ into a base dataset $\mathcal{D}_b$ and a novel dataset $\mathcal{D}_n$, with abundant instances of base classes $\mathcal{C}_b$ and a limited number of instances of novel classes $\mathcal{C}_n$ (i.e., $\mathcal{C}_b \cap \mathcal{C}_n=\emptyset$), respectively. Each input image $x \in \mathcal{X}$ is associated with an annotation $y \in \mathcal{Y}$ comprising the class label $c_i$ and the corresponding bounding box coordinates $b_i$ for each instance $i$. Formally, $\mathcal{D}_b=\{\ (x, y)\ |~y = \{(c_{i},b_{i})\}, c_{i} \in \mathcal{C}_{b}\}$, and $\mathcal{D}_n=\{\ (x, y)\ |~y = \{(c_{i},b_{i})\}, c_{i} \in \mathcal{C}_{n}\}$. 

The training of a G-FSOD consists of two main stages. Firstly, the base training stage strives to build a strong knowledge prior by training on $\mathcal{D}_b$. Secondly, novel training leverages the acquired knowledge to learn the novel classes from $\mathcal{D}_n$ rapidly. Unlike FSOD, G-FSOD aims to maintain its ability to detect $\mathcal{C}_b$ while learning $\mathcal{C}_n$ by leveraging base data samples. G-FSOD performance is measured by the overall Average Precision (AP), which is calculated as a weighted average of the base classes AP (bAP) and novel classes AP (nAP):
\begin{equation}
\small
AP = \frac{|\mathcal{C}_b| \cdot bAP + |\mathcal{C}_n| \cdot nAP}{|\mathcal{C}_b| + |\mathcal{C}_n|}    
\end{equation}
The goal of G-FSOD is to maximize the AP. However, we raise the following arguments: (1) Since the base data samples utilized during novel training are fixed $K$-shots, they do not represent the full base distribution, (2) It is not always possible to store base data samples due to privacy and/or memory constraints. While recent attempts have been made to tackle the former problem~\cite{retentive, cfa}, the latter has not been investigated. 

\subsection{Revisiting Data-Free Knowledge Distillation} 
DFKD approaches aim to distill and transfer knowledge from a teacher to a student network by synthesizing images as an alternative to the old tasks' data. The most  common approaches reviewed in~\cref{sec:related_works} are based on a two-step noise optimization paradigm. First, a noise vector is sampled from a Gaussian distribution and iteratively optimized into a synthesized image with stochastic gradient descent (SGD). This is realized by minimizing the Kullback–Leibler (KL) divergence between the gathered statistics and the statistics yielded by the synthetic images under a Gaussian assumption. The second stage employs standard data-driven knowledge distillation approaches using the synthetic images from the first stage. The aim is to transfer knowledge in a teacher-student fashion to alleviate forgetting.  

To this point, we highlight two main challenges. Firstly, how can we accomplish DFKD in G-FSOD using BN statistics while the majority of the backbone and all BN layers are frozen? Unfreezing such layers during finetuning on the base and novel datasets would lead to parameters and batch statistics that are presumably different from the pre-trained ImageNet ones. As a result, the network will learn new parameters and statistics that differ from the more mature pre-trained ones, devaluing the overall detection performance. Moreover, the BN statistics do not depict the true class-wise means and variances. Secondly, different from image classification, multiple instances per image and the RPN make it challenging to invert the model prior to the RoI head. Otherwise, generating bounding boxes would be needed for the synthesized images incurring higher complexity and significantly more computational and memory overhead.  

\subsection{Stage I: Feature Generator Training}
\label{subsec:generator_training}
Our NIFF framework generates instance-level base features to be replayed during novel class learning. In the first stage, we train a standalone feature generator by aligning the class-wise means and variances at the RoI head. In the second stage, we synthesize features from the generator during novel training to alleviate forgetting and make careful considerations in the training of the pipeline to regularize the detector. In this section, we describe the first stage of our approach addressing the generator design and training.


\textbf{Gathering base statistics.} Since the BN layers are frozen in FSOD and G-FSOD models, an alternative way is needed to gather meaningful statistics (running means and variances) of the base RoI features. Differently from the discussed DFKD methods, we choose to record the class-wise statistics as opposed to class agnostic statistics. This design choice allows more fine-grained control over the number and type of the generated features. A class conditional generation can compensate for the sparser and harder classes in the base dataset.  To this end, we introduce the \textit{data watcher} block which performs average pooling on the spatial dimensions of the input feature maps and records the class-wise mean $\bm{\mu_c}$ and variance $\bm{\sigma_c}^2$ vectors and the sample size $n_c$. Formally, we update the statistics via combined mean and corrected variance as follows:
\begin{equation}
\bm{\hat{\mu}_{c,t}} = \frac{\hat{n}_{c,t-1}\bm{\hat{\mu}_{c,t-1}} + n_{c,t}\bm{\mu_{c,t}}}{\hat{n}_{c,t-1} + n_{c,t}},   
\end{equation}
\begin{equation}
\begin{split}
\bm{\hat{\sigma}_{c,t}}^2 = \frac{(\hat{n}_{c,t-1}-1)\bm{\hat{\sigma}_{c,t-1}}^2 + (n_{c,t}-1)\bm{\sigma_{c,t}}^2}{\hat{n}_{c,t-1} + n_{c,t} - 1}\\
+ \frac{ \hat{n}_{c,t-1} n_{c,t} (\bm{\mu_{c,t}} -\bm{\hat{\mu}_{c,t-1}} )}{(\hat{n}_{c,t-1} + n_{c,t})(\hat{n}_{c,t-1} + n_{c,t} - 1)},     
\end{split}
\end{equation}
where $t$ denotes the iteration step. $\hat{(.)}$ denotes a running estimate. To restrict the generator to forge more diverse features, we opt to place the data watchers at various layers in the RoI head (i.e., before the frozen BN layers and after the activations), as shown in~\cref{fig:stats} (left). The more data watchers we place, the more restricted the forged features are. For this, we further place data watchers before and after the softmax. The collected statistics should be able to depict a strong prior distribution of the base RoI features. It is essential to highlight that once the statistics are collected for the base task, the data is no longer seen or stored. This means that the model enclosing the underlying task statistics can be treated as a black box maintaining data privacy, allowing to share the model with different parties without sharing or storing the data. Note that relying solely on the running averages of the RoI stats results in losing information, especially since we only use RoI pooled features (1024x7x7 fixed output), making it hard to reconstruct the training data. NIFF requires even fewer stats than previous methods like DIODE~\cite{DIODE} (which uses backbone stats) and hence cannot reconstruct an entire image. 




\textbf{Generator architecture.}
Next, we leverage the gathered means and variances of the RoI features to train the proposed feature generator with SGD.   As shown in~\cref{fig:stats} (right), our proposed light-weight generator architecture consists of a linear layer to map the input noise vector $\bm{z} \in \mathbb{R}^{100}$ into $\mathbb{R}^{392}$ which are then reshaped to $\mathbb{R}^{8\times7\times7}$ and fed to $5$ sequential convolutional blocks. Each block comprises of Conv2D layers with $8$ channels and kernel size $3$. Finally, we append a number of $|\mathcal{C}_b|$ $1\times1$ convolutional blocks in parallel to enable generating class-wise instance features $\bm{f_c} \in \mathbb{R}^{1024\times7\times7}$. After sampling noise $\textbf{z} \sim \mathcal{N}(\textbf{0},\textbf{I})$ we generate synthetic features for class $c$ via  $\bm{f_c} = G(c,\bm{z})$, where $G$ is our generator model.

\textbf{Generator training by aligning class-wise statistics.} 
The generator is trained to forge instance-level base features by aligning the acquired statistics in the RoI head at the same layers. The statistics alignment forces the generator to produce diverse base features. To force the generator to produce class-specific features $\bm{f_c}$ from the different heads, we align the class-wise statistics obtained by passing  $\bm{f_c}$ to the RoI head with the class-wise statistics gathered by the data watchers. This is opposed to aligning the class-agnostic full dataset statistics. To further steer each separate head to produce different $f_c$, we add a cross-entropy loss between the corresponding target class-label $\bm{y_{i,c}}$ and the probability $\bm{p_{i,c}}$ at the final softmax layer. 

In summary, the generator is trained using two main objectives: (1) Align the RoI head statistics with the gathered base ones via KL divergence under a Gaussian assumption, and (2) Maximize the class probability via cross-entropy loss. The generator loss function $\mathcal{L}_{\textrm{Gen}}$ is denoted by:\vspace{-0.5em}
 \begin{figure}[t!]
 \centering
 \includegraphics[width=0.80\linewidth]{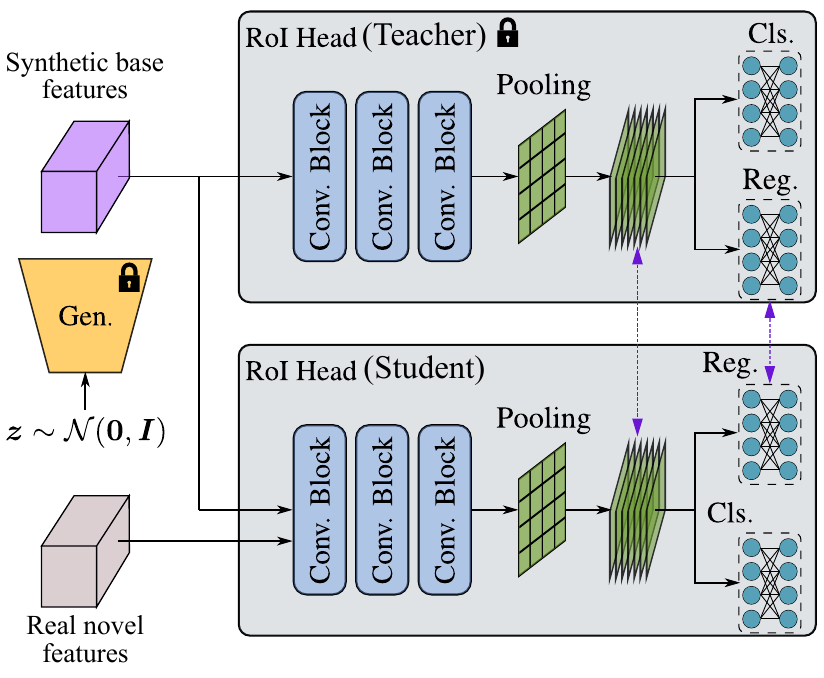}
 \vspace{-1em}
\caption{\textbf{Stage II: Improved Novel Training Pipeline.} During novel training, we perform knowledge distillation via the the trained base feature generator at the RoI head. \vspace{-2em}}
 \label{fig:method}
\end{figure}

{\scriptsize
\begin{align}
\label{eq:kl}
    \nonumber
    \mathcal{L}_{\textrm{Gen}} =& \lambda_{KL} \frac{1}{\abs{\mathcal{C}_b}*d}\sum_{c=1}^{\abs{\mathcal{C}_b}}\sum_{i=1}^{d}\log \frac{\bm{\Tilde{\sigma}_{c,i}}}{\bm{\sigma_{c,i}}} - \frac{1}{2} \Bigg[1 - \frac{\bm{\sigma_{c,i}}^2 + (\bm{\Tilde{\mu}_{c,i}} - \bm{\mu_{c,i}})^2}{\bm{\Tilde{\sigma}_{c,i}}^2}\Bigg]\\
    & - \frac{1}{\abs{\mathcal{C}_b}*N}\sum_{i=1}^{\abs{\mathcal{C}_b}*N} \frac{1}{\abs{\mathcal{C}_b}}\sum_{c=1}^{\abs{\mathcal{C}_b}}\bm{y_{i,c}}\log(\bm{p_{i,c}}).  
\end{align}
}
where $\lambda_{KL}$ is a weighting factor. This loss is averaged over all Data Watchers  (omitted for better readability). $\bm{y_{i}}$ is a one-hot ground-truth vector. $d$ is the pooled feature dimension. $\bm{\mu_{c}}$ and $\bm{\sigma_{c}^2}$ are the gathered feature statistics, while $\bm{\tilde{\mu}_{c}}$ and $\bm{\tilde{\sigma}_{c}^2}$ are the fake features statistics. $N$ is the total number of generated features per class. During training, we set $N=600$. To avoid memory overflow during training, we feed $N$ features for each class sequentially, accumulate the gradients and backpropagate once at the end.

\subsection{Stage II: Improved Novel Training Pipeline}
The final step of NIFF is to finetune on  novel data while performing knowledge distillation at the RoI head in a teacher-student fashion. As depicted in~\cref{fig:method}, along with the real novel features we additionally feed forged base instance-level features via the trained generator network. To match the finetuning $K$-shot setting, we set $N=K$ features per class, which means that all base classes are encountered throughout each iteration. This presents an important advantage in comparison to the data-dependent approaches like~\cite{cfa, retentive} which finetune with base images that contain a few classes only in each iteration. Another advantage is that our approach is generative: this means we can sample from $z$ to produce diverse features, whereas in \cite{cfa, retentive} a fixed number of shots for each base class is presented during training limiting the seen distribution.    

The distillation is performed as follows: First, we employ a weighted feature distillation using L2-norm to penalize the difference between class-wise pooled RoI features of the teacher $\bm{F^T} \in \mathbb{R}^{(\abs{\mathcal{C}_b} * K)\times d}$ and student $\bm{F^S}\in \mathbb{R}^{(\abs{\mathcal{C}_b} * K)\times d}$, where $d$ is the pooled feature dimension. 
Since we generate class-wise features, we weight the difference between the features with the weight vector $\bm{W_\textrm{Cls}^c}\in \mathbb{R}^{d}$ of the entire classification weight matrix $\bm{W_\textrm{Cls}}\in \mathbb{R}^{\abs{\mathcal{C}_b} \times d}$ for the corresponding class $c$. We also weight the regression feature differences in the same way with the corresponding weight matrix $\bm{W_\textrm{Reg}^c}\in \mathbb{R}^{4 \times d}$ from the regression weight tensor $\bm{W_\textrm{Reg}} \in \mathbb{R}^{(4*\abs{\mathcal{C}_b}) \times d}$. Secondly, we align the class-wise regression logits by penalizing the drift between the predicted offsets of the teacher $\bm{reg^T_c}\in \mathbb{R}^4$ and student $\bm{reg^S_c}\in \mathbb{R}^4$. The  distillation loss can be written as:\vspace{-0.5em}
\begin{equation}
\small
\begin{split}
    \mathcal{L}_\textrm{KD} = \lambda_F\frac{1}{\abs{\mathcal{C}_b}*K}\sum_{i=1}^{\abs{\mathcal{C}_b}*K}\norm{(\bm{F^T_i} - \bm{F^S_i}) *   \bm{W_\textrm{Cls}^{c_i}}}_2^2\\ 
    +\lambda_F\frac{1}{4*\abs{\mathcal{C}_b}*K}\sum_{i=1}^{\abs{\mathcal{C}_b}*K}\sum_{j=1}^{4}\norm{(\bm{F^T_i} - \bm{F^S_i}) *   \bm{W_\textrm{Reg}^{c_{i},j}}}_2^2\\ 
    +\frac{1}{\abs{\mathcal{C}_b}*K}\sum_{i=1}^{\abs{\mathcal{C}_b}*K}\norm{\bm{reg^T_i} - \bm{reg^S_i}}_1,
\end{split}
\end{equation}
where $\lambda_F$ is a weighting factor for the feature distillation. Thirdly, similar to~\cite{ABD}, we employ a cross-entropy loss during finetuning to maximize the forged features confidence: \vspace{-0.5em}
\begin{equation}
      \mathcal{L}_\textrm{conf} =  - \frac{1}{\abs{\mathcal{C}_b}*K}\sum_{i=1}^{\abs{\mathcal{C}_b}*K} \frac{1}{\abs{\mathcal{C}}}\sum_{c=1}^{\abs{\mathcal{C}}}\bm{y_{i,c}}\log(\bm{p_{i,c}}).
\end{equation}
 
The overall novel training loss $\mathcal{L}_\textrm{N}$ can be denoted by:
\begin{equation}
    \mathcal{L}_\textrm{N} = \mathcal{L}_\textrm{Cls} + \mathcal{L}_\textrm{Reg} + \mathcal{L}_\textrm{KD} + \mathcal{L}_\textrm{conf} , 
\end{equation}
where $\mathcal{L}_\textrm{Cls}$ and $\mathcal{L}_\textrm{Reg}$ denote the cross-entropy and smooth L1 losses, respectively~\cite{FasterR-CNN}.

\textbf{Additional regularization.} To this point, we find that by replaying the generated base features, the proposed model is almost on par with the state-of-the-art in the overall AP. During novel training, we find that there are important design choices in the training pipeline which can boost the overall detection performance. We opt to perform the recently proposed constraint finetuning approach (CFA)~\cite{cfa}. With the forged base features available, we are able to leverage the backpropagated base gradients to apply CFA. Additionally, we investigate various pixel-level and parameter-level regularization techniques. For the former, we utilize random color jittering (i.e., brightness, contrast, and saturation), random flipping, and random cropping. All augmentations are applied with a probability $p_\textrm{aug} = 0.5$. 

For parameter-level regularization, we find that employing the elastic weight consolidation (EWC)~\cite{ewc} approach, which was originally designed for image classification, can alleviate forgetting. EWC weights the importance of the parameters by computing the diagonal of the Fisher information matrix (FIM). We compute the FIM by squaring the backpropagated gradients during the last epoch in the base training. During the novel training, EWC penalizes the change of important parameters dictated by FIM to allow for a more effective knowledge transfer. However, the FIM occupies a large memory space as it saves a weight for each model parameter. To mitigate this undesirable effect, we average each layer's weights in the FIM, reducing the memory from $\sim 200\textrm{MB}$ to $6.8\textrm{KB}$. Thus, each layer's importance is represented by a single scalar value. We denote this approximation of EWC by mean EWC (mEWC). We show in the supplementary that mEWC is on par with EWC when applied to our model.

\section{Experiments}

We evaluate our proposed approach using the well-established G-FSOD benchmarks, including MS-COCO~\cite{coco} and PASCAL-VOC~\cite{pascalvoc} datasets. We employ the same data splits as in earlier works~\cite{TFA, defrcn, cfa} to ensure a fair comparison. Implementation details are provided in the Supplementary material.

\textbf{Datasets.} Firstly, the MS-COCO dataset contains 80 classes, of which 60 are base categories disjoint with PASCAL-VOC, and the remaining 20 are unique classes. We use $5k$ validation set during testing, while the rest is used for training. We report the outcomes of $K=5,10,30-$shot settings.
Secondly, the PASCAL-VOC dataset has three distinct splits, each with 20 classes. Further, the classes are randomly divided into 15 and 5, respectively, base and novel classes. For base and novel training, data is drawn from the VOC 2007 and VOC 2012 train/val sets. The VOC 2007 test set is used for testing. The outcomes of $K=1,2,3,5,10$-shot settings are reported. 

\textbf{Evaluation metrics.} Following prior G-FSOD works~\cite{retentive,cfa}, we report the overall (AP), base (bAP), and novel (nAP) metrics. We also compute the Average Recall (AR) for the base (bAR) and novel (nAR) classes. Finally, we report the ensemble-inference results (w/E), leveraging the base model parameters during inference~\cite{retentive,cfa}.

\subsection{Ablation Studies}
In order to showcase the importance of our novel contributions and design choices, we ablate different parts of the pipeline on the MS-COCO dataset under the $10$-shot setting.

\begin{table}[t!]
\begin{center}
\scalebox{0.75}{
\begin{tabular}{l|ccc|c|c}
	\toprule
  \multirow{2}{*}{Model Configuration} & \multicolumn{5}{c}{10-Shot Inference}\\
  &  AP & bAP & nAP & AR & Memory [MB]\\
\midrule
Inverted RoI head & 28.9 & 32.6 & 17.7 & 27.1 & \textbf{0}\\
\midrule
Gen. w/ shared head  &\multirow{2}{*}{29.3} & \multirow{2}{*}{33.1} & \multirow{2}{*}{17.8} & \multirow{2}{*}{27.3} & \multirow{2}{*}{1.6} \\
w/o classwise stats  & & & & &  \\
\midrule
Gen. w/ shared head & \multirow{2}{*}{28.5} & \multirow{2}{*}{32.1} & \multirow{2}{*}{17.7} & \multirow{2}{*}{26.4} & \multirow{2}{*}{1.6} \\
w/ classwise stats & & & & \\
\midrule
Gen. w/ separate heads & \multirow{2}{*}{29.0} & \multirow{2}{*}{32.8} & \multirow{2}{*}{17.5} & \multirow{2}{*}{27.4} & \multirow{2}{*}{3.7} \\
w/o classwise stats & & & & \\
\midrule
Gen. w/ separate heads & \multirow{2}{*}{\textbf{30.7}} & \multirow{2}{*}{\textbf{35.0}} & \multirow{2}{*}{17.8} & \multirow{2}{*}{28.6} & \multirow{2}{*}{3.7}\\
w/ classwise stats \textbf{(Ours)} & & & & \\
\midrule
Ours w/o~cross-entropy term & 30.0 & 34.1 & 17.6 & 28.6 & 3.7\\
\midrule
Ours w/~($\textrm{dim}=64$) & \textbf{30.7} & \textbf{35.0} & 17.8 & \textbf{28.8} & 28.7\\
Ours w/~($\textrm{dim}=32$) & 30.5 & 34.8 & \textbf{17.9} & 28.6 & 14.0\\
Ours w/~($\textrm{dim}=16$) & \textbf{30.7} & 34.9 & \textbf{17.9} & 28.7 & 7.1\\
Ours w/~($\textrm{dim}=8$) & \textbf{30.7} & \textbf{35.0} & 17.8 & 28.6 & 3.7\\

\bottomrule
\end{tabular}}
\end{center}
\vspace{-1.5em}
\caption{Impact of various generator design choices on MS-COCO ($10$-shot) without the added regularization. By memory, we mean the overhead storage needed other than the detection model.}
\label{tab:gen_design}
\vspace{-1.5em}
\end{table}
\begin{table*}
\small
\setlength{\tabcolsep}{10pt}
    \centering
    \resizebox{0.9\textwidth}{!}{\begin{tabular}{c | c | c | c c c | c c c | c c c}
      \Xhline{1pt}
      \multirow{2}{*}{\textbf{Methods} / \textbf{Shots}} & \multirow{2}{*}{\textbf{w/E}} & \multirow{2}{*}{\textbf{w/B}} &
      \multicolumn{3}{c |}{\textbf{5 shot}}  &
      \multicolumn{3}{c |}{\textbf{10 shot}}  &
      \multicolumn{3}{c}{\textbf{30 shot}}\\
      & & & AP &  bAP & nAP & AP &  bAP & nAP & AP & bAP & nAP \\ \Xhline{1pt}
      FRCN-ft-full\cite{TFA} & \xmark & \cmark & 18.0 & 22.0 & 6.0 & 18.1 & 21.0 & 9.2 & 18.6 & 20.6 & 12.5 \\
      TFA w/ fc\cite{TFA} & \xmark & \cmark & 27.5 & 33.9 & 8.4 & 27.9 & 33.9 & 10.0 & 29.7 & 35.1 & 13.4 \\
      TFA w/ cos\cite{TFA} & \xmark & \cmark & 28.1 & 34.7 & 8.3 & 28.7 & 35.0 & 10.0 & 30.3 & 35.8 & 13.7 \\
      MPSR\cite{MPSR} & \xmark & \cmark & - & - & - & 15.3 & 17.1 & 9.7 & 17.1 & 18.1 & 14.1 \\
      DeFRCN~\cite{defrcn} & \xmark & \cmark & 28.7 & 33.1 & 15.3 & 30.6 & 34.6 & \second{18.6} & 31.6 & 34.7 & \second{22.5}\\
      ONCE~\cite{once} & \xmark & \cmark & 13.7 & 17.9 & 1.0 & 13.7 & 17.9 & 1.2 & - & - & -  \\
      Meta R-CNN~\cite{MetaRCNN}& \xmark & \cmark & 3.6 & 3.5 & 3.8 & 5.4 & 5.2 & 6.1 & 7.8 & 7.1 & 9.9 \\
      FSRW\cite{FSRW}& \xmark & \cmark & - & - & - & - & - & 5.6 & - & - & 9.1 \\
      FsDetView~\cite{FsDetView}& \xmark & \cmark & 5.9 & 5.7 & 6.6 & 6.7 & 6.4 & 7.6 & 10.0 & 9.3 & 12.0 \\
      CFA w/ fc~\cite{cfa} & \xmark & \cmark & \second{30.1} & \best{37.1} & 9.0 & 30.8 & \best{37.6} & 10.5 & 31.9 & \best{37.7} & 14.7\\
      CFA w/ cos~\cite{cfa}& \xmark & \cmark & 29.7 & \second{36.3} & 9.8 & 30.3 & \second{36.6} & 11.3 & 31.7 & 37.0 & 15.6 \\
      CFA-DeFRCN~\cite{cfa}& \xmark & \cmark & \second{30.1} & 35.0 & \second{15.6} & \second{31.4} & 35.5 & \best{19.1} & \second{32.0} & 35.0 & \best{23.0} \\
      \Xhline{1pt}
      DeFRCN & \xmark & \xmark & 23.7 & 26.3 & \second{15.6} & 18.2 & 18.5 & 17.4 & 16.3 & 16.3 & 21.4\\
      \rowcolor[HTML]{EFEFEF}
      NIFF-DeFRCN & \xmark & \xmark & \best{31.3} & \second{36.3} & \best{15.7} & \best{32.2} & \second{36.6} & \best{19.1} & \best{33.1} & \second{37.2} & 21.0 \\
      \midrule[1.5pt]
      Retentive R-CNN\cite{retentive}& \cmark & \cmark& 31.5 & \second{39.2} & 8.3 & 32.1 & 39.2 & 10.5 & 32.9 & \second{39.3} & 13.8 \\
      CFA w/ fc~\cite{cfa} & \cmark & \cmark& 31.8 & \best{39.5} & 8.8 & 32.2 & \best{39.5} & 10.4 & 33.2 & \best{39.5} & 14.3\\
      CFA w/ cos~\cite{cfa}& \cmark & \cmark& 32.0 & \best{39.5} & \second{9.6} & \second{32.4} & \second{39.4} & 11.3 & 33.4 & \best{39.5} & 15.1 \\
      CFA-DeFRCN~\cite{cfa}& \cmark & \cmark& \second{33.0} & 38.9 & \second{15.6} & \best{34.0} & 39.0 & \best{18.9} & \best{34.9} & 39.0 & \best{22.6} \\
      \Xhline{1pt}
      \rowcolor[HTML]{EFEFEF}
      NIFF-DeFRCN &\cmark  & \xmark& \best{33.1} & 38.9 & \best{15.9} & \best{34.0} & 39.0 & \second{18.8} & \second{34.5} & 39.0 & \second{20.9} \\
      \Xhline{1pt}
    \end{tabular}}
    \vspace{-0.5em}
    \caption{G-FSOD results on MS-COCO for $5,10,30$-shot settings. w/E denotes the ensemble-based evaluation protocol. w/B indicates whether the base data is available during novel finetuning. The {\color{red}best} and {\color{blue}second-best} results for each evaluation protocol are color coded. \vspace{-1.5em}}
    \label{tab:coco}
\end{table*}

\textbf{Generator design choices.} In \cref{tab:gen_design}, we validate our generator design choices without any regularization (namely: \textit{standalone generator}, \textit{class-wise statistics}, \textit{separate heads} and \textit{number of channels per layer}). First, we invert the RoI head to produce instance-level features by minimizing a KL-loss with respect to the gathered base data statistics. Note that this can be considered as a straightforward extension of the standard MI approach ABD~\cite{ABD} to G-FSOD, albeit this model produces features not images. Next, we train a standalone generator with a shared head for all classes while minimizing the full base-data statistics. Although the generator adds a negligible memory overhead, it outperforms the inverted model, which supports our claim that a separate generator is easier to optimize. However, when trained with class-wise statistics the performance slightly drops. We then replace the shared head with separate class-aware heads and and minimize the full base statistics. We notice that this setting is on par with the generator w/ shared head in row $2$. Only when we combine the separate heads with the class-wise statistics can we achieve the best overall performance, as the model can now better consider the inter-class variance. Solely extending the model by class-wise statistics or by class-wise heads reduces the overall performance.  Afterwards, we experiment with the complete version of the generator but we remove the cross-entropy loss from \cref{eq:kl} and observe a slight drop in the overall AP. In the lower part of the table, we study the trade-off between AP and memory by changing the number of channels per generator layer. Interestingly, we find that the model with 64 channels and the model with 8 channels perform similarly, so we choose the minimalist design to reduce the memory footprint.

\begin{table}[t!]
\begin{center}
\scalebox{0.85}{
\begin{tabular}{l|ccc|ccc}
	\toprule
  \multirow{2}{*}{Feature(s) per class} & \multicolumn{6}{c}{10-Shot Inference}\\
  &  AP & bAP & nAP & AR & bAR & nAR\\
  \midrule
1  & 29.9 & 34.0 & 17.6 & 28.4 & 31.0 & \second{20.7} \\  
5  & 30.6 & 34.9 & \second{17.7 }                     & \best{28.8} &\best{31.5} &\best{20.8}\\  
10 & \second{30.7} & \second{35.0} & \best{17.8}               & \best{28.8} &\best{31.5} &\best{20.8}\\  
30 & \best{30.8} & \best{35.1} & \best{17.8} & \best{28.8} &\best{31.5} &\best{20.8}\\   
\midrule
10 (fixed)& 25.9 & 28.9 & 16.9 & 25.5 & 27.1 & 20.6\\ 
\midrule
10 (10 sampled cls)& 30.5 & 34.8& \second{17.7} & \second{28.7} & \second{31.4} & 20.6\\
\bottomrule
\end{tabular}}
\end{center}
\vspace{-1.5em}
\caption{Effect of the number of generated features per class and fixed samples on the forgetting and the detection performance without any added regularization.}
\label{tab:ablation_features}
\vspace{-2em}
\end{table}
\textbf{Impact of generated features sampling.} In~\cref{tab:ablation_features}, we study the impact of sampling techniques when generating features during novel finetuning (no regularization is performed). We start by generating only $1$ feature per class, then we experiment with a higher number of features to verify its effect on the overall AP. We observe that once we reach $10$ features per class which matches the novel finetuning setting ($N=K=10$-shot), the best overall results are achieved. Further increases yield almost similar results so we opt to set $N=K$ consistently throughout our experiments on MS-COCO and PASCAL-VOC. Next, in row ($5$), we generate $10$ features per class by sampling only once at the start of the training and fix them throughout the novel training. We notice that the performance drops significantly due to the limited diversity of generated base features, emphasizing the importance of sampling the features in each iteration. In the final row, we randomly sample a random subset of the base classes $\mathcal{C}_s < |\mathcal{C}_b|$, but still generate $N=10$ features for each. We choose $\mathcal{C}_s = 10$. Compared to generating features for all the base classes (row $3$), we notice a slight drop in performance, highlighting the importance of a class-balanced sampling scheme. Hence, we opt to generate $N=K$ features per class for all the base classes in each iteration to achieve the best overall AP.      
\begin{table}[t!]
\small
\setlength{\tabcolsep}{3pt}
    \centering
    \adjustbox{width=\linewidth}{
    \begin{tabular}{l | c | c c c | c c c | c c c}
      \Xhline{1pt}
      \multirow{2}{*}{\textbf{Methods} / \textbf{Shots}} & \multirow{2}{*}{\textbf{w/B}} &
      \multicolumn{3}{c |}{\textbf{5-Shot}}  &
      \multicolumn{3}{c |}{\textbf{10-Shot}}  &
      \multicolumn{3}{c}{\textbf{30-Shot}}\\
      & & AP &  bAP & nAP & AP &  bAP & nAP & AP & bAP & nAP \\ \Xhline{1pt}
      DeFRCN~\cite{defrcn} & \cmark & 28.7 & 33.1 & 15.3 & 30.6 & 34.6 & \second{18.6} & 31.6 & 34.7 & \best{22.5}\\
      \Xhline{1pt}
      DeFRCN         & \xmark  & 23.7 & 26.3 & \second{15.6} & 18.2 & 18.5 & 17.4 & 16.3 & 16.3 & 21.4\\
      DeFRCN w/ DA   & \xmark  & 22.6 & 25.0 & 15.3          & 26.4 & 29.2 & 17.9 & 24.2 & 25.0 & \second{21.8}\\
      DeFRCN w/ LOD  & \xmark  & 29.0 & 34.1 & 13.4          & 27.0 & 30.7 & 16.1 & 29.8 & 33.2 & 19.7\\
      DeFRCN w/ MAS  & \xmark  & 31.0 & \second{36.8} & 13.5          & 31.5 & \second{36.8} & 15.3 & 32.6 & 36.6 & 20.4\\
      DeFRCN w/ EWC  & \xmark  & \second{31.1} & \best{37.1} & 13.4          & \second{31.8} & \best{36.9} & 16.6 & \second{33.0} & \best{37.3} & 20.1\\
      \Xhline{1pt}
       DeFRCN & \cmark$_F$ & 24.2 & 27.6 & 13.6 & 25.8 & 28.9 & 16.6 & 26.6 & 29.0 & 19.7\\
       DeFRCN + CFA & \cmark$_F$ & 26.0 & 29.9 & 13.7& 27.7 & 31.4 & 16.6 & 28.6 & 31.5 & 19.9\\
       DeFRCN w/ KD & \cmark$_F$ & 25.3 & 28.8 & 14.3& 27.0 & 30.2 & 17.4 & 27.9 & 30.3 & 20.5\\
       DeFRCN w/ KD + CFA & \cmark$_F$ & 28.4 & 33.3 & 14.1 & 30.5 & 34.9 & 17.1 & 31.3 & 35.0 & 20.3\\      
      \Xhline{1pt}
      \rowcolor[HTML]{EFEFEF}
      NIFF-DeFRCN & \xmark & \best{31.3} & 36.3 & \best{15.7} & \best{32.2} & 36.6 & \best{19.1} & \best{33.1} & \second{37.2} & 21.0 \\
      \bottomrule
    \end{tabular}}
    \vspace{-0.5em}
    \caption{G-FSOD results for various baselines on top of DeFRCN~\cite{defrcn} on MS-COCO for $5,10,30$-shot settings. \cmark$_F$ denotes finetuning with offline saved base RoI features.
    w/B indicates whether the base data is available during novel finetuning. KD denotes the proposed knowledge distillation approach.\vspace{-2em}}
    
    \label{tab:main_baselines}
\end{table}

\begin{table*}[t!]
\scriptsize
    \centering
    \begin{tabular}{c| c | c| c c c c c | c c c c c | c c c c c}
      \Xhline{1pt}
      \multirow{2}{*}{\textbf{Methods} / \textbf{Shots}} &
      \multirow{2}{*}{\textbf{w/E}} &
      \multirow{2}{*}{\textbf{w/B}} &
      \multicolumn{5}{c |}{\textbf{All Set 1}}  &
      \multicolumn{5}{c |}{\textbf{All Set 2}}  &
      \multicolumn{5}{c}{\textbf{All Set 3}}\\

      & & & 1 &  {2} & {3} & {5} &  {10} & 1 &  {2} & {3} & {5} &  {10} & 1 &  {2} & {3} & {5} &  {10} \\ \Xhline{1pt}
      FRCN-ft-full\cite{TFA} &\xmark &\cmark& 55.4 & 57.1 & 56.8 & 60.1 & 60.9 & 50.1 & 53.7 & 53.6 & 55.9 & 55.5 & 58.5 & 59.1 & 58.7 & 61.8 & 60.8 \\
      TFA w/ fc\cite{TFA}&\xmark &\cmark& 69.3 & 66.9 & 70.3 & 73.4 & 73.2 & 64.7 & 66.3 & 67.7 & 68.3 & 68.7 & 67.8 & 68.9 & 70.8 & 72.3 & 72.2 \\
      TFA w/ cos\cite{TFA}&\xmark &\cmark& 69.7 & 68.2 & 70.5 & 73.4 & 72.8 & 65.5 & 65.0 & 67.7 & 68.0 & 68.6 & 67.9 & 68.6 & 71.0 & 72.5 & 72.4 \\
      MPSR\cite{MPSR}&\xmark &\cmark& 56.8 & 60.4 & 62.8 & 66.1 & 69.0 &
       53.1 & 57.6 & 62.8 & 64.2 & 66.3 & 55.2 & 59.8 & 62.7 & 66.9 & 67.7 \\
      DeFRCN\cite{defrcn}&\xmark &\cmark & 73.1 & 73.2 & 73.7 & 75.1 & \second{74.4} & 68.6 & \second{69.8} & 71.0 & 72.5 & 71.5 & 72.5 & 73.5 & 72.7 & 74.1 & 73.9 \\
       Meta R-CNN\cite{MetaRCNN}&\xmark &\cmark & 17.5 & 30.5 & 36.2 & 49.3 & 55.6 & 19.4 & 33.2 & 34.8 & 44.4 & 53.9 & 20.3 & 31.0 & 41.2 & 48.0 & 55.1 \\
       FSRW\cite{FSRW}&\xmark &\cmark & 53.5 & 50.2 & 55.3 & 56.0 & 59.5 & 55.1 & 54.2 & 55.2 & 57.5 & 58.9 & 54.2 & 53.5 & 54.7 & 58.6 & 57.6 \\
       FsDetView\cite{FsDetView}&\xmark &\cmark & 36.4 & 40.3 & 40.1 & 50.0 & 55.3 & 36.3 & 43.7 & 41.6 & 45.8 & 54.1 & 37.0 & 39.5 & 40.7 & 50.7 & 54.8 \\ 
        CFA w/ fc~\cite{cfa}&\xmark &\cmark & 69.5 & 68.2 & 69.8 & 73.5 & 74.3  & 66.0 & 66.9 & 69.2 & 70.1 & 71.1 & 67.7 & 69.0 & 70.9 & 72.6 & 73.5 \\
        CFA w/ cos~\cite{cfa}&\xmark &\cmark & 69.1 & 69.8 & 71.9 & 73.6 & 73.9 & 64.8 & 66.5 & 68.3 & 69.5 & 70.5 & 67.7 & 69.7 & 71.9 & 73.0 & 73.5 \\
        CFA-DeFRCN~\cite{cfa}&\xmark &\cmark & \second{73.8} & \second{74.6} & \second{74.5} & \second{76.0} & \second{74.4} & \second{69.3} & \best{71.4} & \second{72.0} & \second{73.3} & \second{72.0} & \second{72.9} & \second{73.9} & \second{73.0} & \second{74.1} & \second{74.6} \\
       \Xhline{1pt}
        DeFRCN&\xmark &\xmark & 61.1 & 48.5 & 35.9 & 32.8 & 20.7 & 64.7 & 59.7 & 58.2 & 56.9 & 48.4 & 56.3 & 51.2 & 46.9 & 38.8 & 23.9\\ 
        \rowcolor[HTML]{EFEFEF}
        NIFF-DeFRCN&\xmark &\xmark & \best{75.6} & \best{76.5} & \best{76.7} & \best{77.4} & \best{76.9} & \best{70.0} & \best{71.4} & \best{73.9} & \best{74.4} & \best{74.0} & \best{74.4} & \best{75.8} & \best{76.2} & \best{76.6} & \best{76.7} \\
        \midrule[1.5pt]
       Retentive R-CNN\cite{retentive}&\cmark &\cmark & 71.3 & 72.3 & 72.1 & 74.0 & 74.6 & 66.8 & 68.4 & 70.2 & 70.7 & 71.5 & 69.0 & 70.9 & 72.3 & 73.9 & 74.1 \\
       CFA w/ fc~\cite{cfa}&\cmark &\cmark & 70.3 & 69.5 & 71.0 & 74.4 & 74.9 & 67.0 & 68.0 & 70.2 & 70.8 & 71.5 &  69.1 & 70.1 & 71.6 & 73.3 & 74.7 \\
       CFA w/ cos~\cite{cfa}&\cmark &\cmark & 71.4 & 71.8 & 73.3 & 74.9 & 75.0 & 66.8 & 68.4 & 70.4 & 71.1 & 71.9 & \second{69.7} & 71.2 & 72.6 & 74.0 & 74.7 \\
       CFA-DeFRCN~\cite{cfa}&\cmark &\cmark & \second{75.0} & \second{76.0} & \second{76.8} & \second{77.3} & \second{77.3} & \second{70.4} & \best{72.7} & \second{73.7} & \second{74.7} & \second{74.2} & \best{74.7} & \second{75.5} & \second{75.0} & \second{76.2} &  \second{76.6} \\ 
       \Xhline{1pt}
        \rowcolor[HTML]{EFEFEF}
        NIFF-DeFRCN&\cmark &\xmark & \best{75.9} & \best{76.9} & \best{77.3} & \best{77.9} & \best{77.5} & \best{70.6} & \second{71.6} & \best{74.5} & \best{75.1} & \best{74.5} & \best{74.7} & \best{76.0} & \best{76.1} & \best{76.8} & \best{76.7} \\ \Xhline{1pt}
    \end{tabular}
    \vspace{-1em}
    \caption{G-FSOD (AP50) results on PASCAL-VOC for $1,2,3,5,10$-shot settings for all three splits are reported. }
    \label{tab:voc-all}
    \vspace{-2.5em}
\end{table*}
\vspace{-0.5em}
\subsection{Main Comparisons}
\vspace{-0.5em}
We compare our method (NIFF) against state-of-the art G-FSOD~\cite{retentive, cfa} and FSOD~\cite{TFA, MPSR,defrcn} models on MS-COCO and PASCAL-VOC benchmarks. We opt to apply our approach on top of the recent state-of-the-art transfer learning based approach DeFRCN~\cite{defrcn}. We denote our model by \textit{NIFF-DeFRCN}.  

\textbf{Results on MS-COCO.} In~\cref{tab:coco}, we show the results on MS-COCO. (w/B) denotes whether base data is used. To show the impact of removing the base data on G-FSOD, we reevaluate our baseline DeFRCN~\cite{defrcn} without any  base data. We notice that both the base and novel performance drop across all shots. This shows the importance of the base data in the knowledge transfer to new tasks. By applying NIFF, we show that we are consistently able to boost the base performance across all settings yielding higher overall AP performance. Moreover, we evaluate our model using the ensemble evaluation protocol in Retentive R-CNN~\cite{retentive} and  outperform the other approaches despite the absence of base data (with an $0.4$AP less in $30$-shot setting). It is essential to note that this evaluation protocol requires keeping the base model parameters~\cite{retentive,cfa}, increasing the overall memory footprint and inference time. Note that our approach NIFF-DeFRCN (w/o E and w/o B) outperforms Retentive R-CNN (w/E and w/B) in the overall AP.    

\textbf{Comparison against regularization-based continual learning and G-FSOD baselines.} In~\cref{tab:main_baselines}, we compare our method with base-data-free and GFSOD baselines on top of DeFRCN~\cite{defrcn}. The data-free baselines are drawn from regularization-based continual learning works, namely: pixel-level data augmentations (DA), EWC~\cite{ewc} and memory aware synapses (MAS)~\cite{mas} with the computed FIM, and lifelong object detection (LOD)~\cite{lod}. Since CFA~\cite{cfa} cannot be conducted in a data-free setting, we train DeFRCN and save base RoI features to later apply CFA during novel training. We also add two baselines: the proposed KD method on top of DeFRCN using saved base RoI features with and without CFA. The proposed method is able to consistently outperform both data-free and data-reliant baselines. We argue that the diversity of the forged features is what allows our method to surpass data-reliant baselines. Moreover, it is important to note that the stored features require $114.8$MB, which is significantly more than the memory required by our generator ($3.7$MB) and stats ($12.4$MB). Similarly, EWC and MAS are memory intensive as the FIM requires around $200$MB.  

\textbf{Results on PASCAL-VOC.} We report the overall performance on PASCAL-VOC (AP50) in~\cref{tab:voc-all}. We show that adopting NIFF achieves state-of-the-art results with and without the ensemble evaluation protocol.  Due to the limited space, we report the novel performance (nAP50) results on PASCAL-VOC. Further ablation experiments on MS-COCO and experiments with multiple runs on both datasets in the Supplementary materials. Although the performance on novel classes is not our primary objective, NIFF-DeFRCN achieves competitive results on both datasets are shown in the majority of cases.
\subsection{Memory \& Computation Analysis}
For 10-shot MS-COCO, the overall memory required is (Model = 195.16MB, Base Images = 148.8MB, Novel Images= 48.6MB). The Generator has 3.7MB, the stats have 12.42MB, so the proposed model saves $33.8\%$ of the initial requirements.  Computationally, DeFRCN and the generator require $133.46$G and $943.94$K FLOPS, respectively, implying that the computational overhead is negligible. DeFRCN G-FSOD training requires $104.5$ mins, where our generator training and data generation additionally require $112$ mins and $27$ mins, respectively. 

\section{Conclusion}
We propose NIFF, a data-free G-FSOD framework, which alleviates forgetting on the base images all without using any base data. Our main contribution is the design of a \textit{standalone} generator that forges base \textit{instance-level features} instead of images by aligning \textit{class-wise statistics} in the RoI head. The generator has a negligible memory footprint ($\sim4$MB) which is two orders of magnitude lower than using base images for finetuning. Replaying the forged features during novel finetuning, along with careful design choices in the training pipeline, results in state-of-the-art overall AP performance. In the future, we plan to apply our approach on the recent state-of-the-art transformer-based FSOD models. Furthermore, we encourage future works to extend G-FSOD in meta-learning paradigms as their base performance drops significantly. Finally, we hope that our proposed approach of using a standalone generator in the feature space sheds light on works in areas other than object detection.

\vspace{0.9em}
\setstretch{0.6}
{\noindent\small \textbf{Acknowledgment} This research was supported by the German Federal Ministry for Economic Affairs and Climate Action (BMWK) within the project FabOS under grant number 01MK20010I. The authors would like to thank Mohamed Sayed from UCL for the helpful and informative discussions.}
\clearpage
{\small
\bibliographystyle{unsrt}
\bibliography{egbib}
}

\end{document}


\title{[Supplementary] NIFF: Alleviating Forgetting in Generalized Few-Shot Object Detection via Neural Instance Feature Forging }  
\maketitle
\appendix

\section{Implementation Details}

\subsection{Generator Training}
Using the base trained RoI head parameters and the gathered statistics in the data watchers, we train the generator for $2$k iterations. We optimize the generator using SGD with a batch size of $600$ features, momentum of $0.9$, and weight decay of $5e^{-5}$. The learning rate is set to $0.001$. The scaling factor for the KL divergence loss is set to $\lambda_\textrm{KL}=5$.  

\subsection{Novel Training}
During novel training, the model is also optimized using SGD with batch size of $16$ and learning rate of $0.005$ and $0.01$ for MS-COCO and PASCAL-VOC, respectively. We set the warmup iterations to $200$. Step decays are performed at $2500$, $4000$, and $6400$ for MS-COCO $5$, $10$,and $30$-shot settings, respectively. As for PASCAL-VOC, we perform the decay for the first three shot settings at $1000$ and $1500$ for the $5$ and $10$-shot settings. To perform the distillation, we unfreeze the RoI head and down scale its learning rate by a factor of $0.015$. We set the scaling factors as follows: $\lambda_\textrm{F}=0.1$ and $0.01$ for the mEWC penalty term.

\section{Further Ablations}
\begin{table}[h!]
\begin{center}
\scalebox{0.85}{
\begin{tabular}{l|ccc|ccc}
	\toprule
  \multirow{2}{*}{Data Watcher Config.} & \multicolumn{6}{c}{10-Shot Inference}\\
  &  AP & bAP & nAP & AR & bAR & nAR\\
  \midrule
(1) After Act.& 32.0 & 36.7 & 18.0 & 29.9 & 32.7 & 21.5 \\   
(2) Before FBN & \textbf{32.2} & \textbf{36.9} & 18.1 & \textbf{30.1} & \textbf{33.0} & 21.1 \\  
\rowcolor[HTML]{EFEFEF}
(3) Both & \textbf{32.2} & 36.6 & \textbf{19.1} & 29.6 & 32.1& \textbf{22.3}\\
\bottomrule
\end{tabular}}
\end{center}
\vspace{-1.5em}
\caption{Where to place the data watchers to record useful statistics for better feature generation with respect to the frozen BatchNorm layers and the activations that follow. Results on MS-COCO 10-shot.\vspace{-1em}}
\label{tab:data_watchers}
\end{table}
\textbf{Which statistics matter?} In \cref{tab:data_watchers}, we study which statistics are needed to capture the base data distribution. To this end, we place data watchers in different places and train different generators accordingly. Then we finetune on the novel data in each setup and report the final detection results. The different locations for placing data watchers are: (1) after the activations (Act.) following the frozen-BN (FBN), (2) before the FBN layers, (3) both locations. As we can observe, locations (2) and (3) yield empirically better results than (1). Although the AP of (2) and (3) is the same, we choose (3) as its nAP is slightly higher. 

\begin{table}[t!]
\begin{center}
\scalebox{0.8}{
\begin{tabular}{ll|ccc|ccc}
	\toprule
  & \multirow{2}{*}{Model Configuration} & \multicolumn{6}{c}{10-Shot Inference}\\
  & &  AP & bAP & nAP & AR & bAR & nAR\\
  \midrule
\textbf{0} &  DeFRCN & 30.6 & 34.6 & 18.6 & 29.1 & 32.0 & 20.5\\ 
\hline
\textbf{A} &  DeFRCN (Base-Free) & 18.2 & 18.5 & 17.4 & 17.5 & 16.2 & 21.3\\  
\textbf{B} &  + Generator & 30.7 & 35.0 & 17.8 & 28.6 & 31.3& 20.9\\  
\textbf{C} & + CFA & 32.0 & 36.4 & 18.5 & 29.6 & 32.4 & 21.3\\  
\textbf{D} & + DA  & \textbf{32.2} & \textbf{36.8} & 18.4 & \textbf{29.9} & \textbf{32.6}& 21.7\\   
\rowcolor[HTML]{EFEFEF}
\textbf{E} & + mEWC (NIFF) &  \textbf{32.2} & 36.6 & \textbf{19.1} & 29.6 & 32.1 & \textbf{22.3}\\  
\bottomrule
\end{tabular}}
\end{center}
\vspace{-1.5em}
\caption{Incremental component analysis on MS-COCO ($10$-shot).\vspace{-1.0em}}
\label{table:ablation_incremental_components}
\end{table}

\textbf{Component analysis.} In \cref{table:ablation_incremental_components}, we start by reporting the results of the state-of-the-art model in transfer-learning namely, DeFRCN~\cite{defrcn} (Config \textbf{0})). In Config \textbf{A}, we train a DeFRCN model on novel data without using the base classes. We consider this as our baseline, to which we gradually introduce our contributions in an incremental fashion. Note that without using base data, DeFRCN performance drops dramatically on the base classes (by $40.5\%$). In Config \textbf{B}, we train our standalone lightweight generator and finetune DeFRCN on novel data while replaying synthetic base features from the generator. We show that we are able to almost recover the overall performance of DeFRCN in Config \textbf{0}, all without using any base data. Next, in Config \textbf{C}, we apply CFA~\cite{cfa} on the gradients which are backpropagated on the RoI head. In Config \textbf{D} and \textbf{E}, we add the chosen pixel-level data augmentation (DA) and parameter-level (mEWC) regularization techniques, which allow us to achieve state-of-the-art results in the overall performance.  

\subsection{Generator Architecture}
\begin{table}[h!]
\begin{center}
\scalebox{0.9}{
\begin{tabular}{l|ccc|c}
	\toprule
  \multirow{2}{*}{Model Configuration} & \multicolumn{4}{c}{10-Shot Inference}\\
  &  AP & bAP & nAP & AR\\
\midrule
Number of Layers (L=3)  & 27.6 & 31.3 & 16.4 & 27.6\\
Number of Layers (L=5)  & 30.7 & 35.0 & 17.7 & 28.8\\
Number of Layers (L=7)  & 31.2 & 35.7 & 17.7 & 29.3\\
Number of Layers (L=10)  & 31.1 & 35.6 & 17.5 & 29.2\\
\midrule
Kernel Size (K=1)  & 30.2 & 34.7 & 17.7 & 28.3\\
Kernel Size (K=3)  & 30.7 & 35.0 & 17.7 & 28.8\\
Kernel Size (K=5)  & 30.5 & 34.8 & 17.6 & 28.7\\
Kernel Size (K=7)  & 30.6 & 34.8 & 17.8 & 28.6\\
\midrule
Noise Dimension (z=50)  & 30.5 & 34.9 & 17.6 & 28.5\\
Noise Dimension (z=100)  & 30.7 & 35.0 & 17.7 & 28.8\\
Noise Dimension (z=1000)  & 30.7 & 35.0 & 17.7 & 28.8\\
\bottomrule
\end{tabular}}
\end{center}
\vspace{-1.5em}
\caption{Impact of various generator architectural design choices on MS-COCO ($10$-shot). We finetune DeFRCN without base data using the generator without any regularization techniques.}
\label{tab:supp_ablation_gen}
\vspace{-1em}
\end{table}
In~\cref{tab:supp_ablation_gen}, we investigate the impact various architectural design choices on the overall detection performance.  The utilized generator throughout this work comprises five convolutional layers ($L=5$) with kernel size ($K=3$) and an input noise dimension ($z=100$).
To study the effect of each of these design choices, we change only one factor and leave the rest unchanged. Firstly, increasing the number of layers contribute to higher overall performance where the base performance is noticeably improved. However, past $L=7$ the performance starts to slightly drop. Secondly, we show that increasing the kernel size do not result in any performance gain. Finally, increasing the noise dimension past $z=100$ yields no change in the performance implying that the generator is able to generate diverse high-quality features without requiring high-dimensional noise vectors. 

\subsection{mEWC VS EWC}
\begin{table}[h!]
\begin{center}
\scalebox{0.9}{
\begin{tabular}{l|ccc|c}
	\toprule
  \multirow{2}{*}{Configuration} & \multicolumn{4}{c}{10-Shot Inference}\\
  &  AP & bAP & nAP & AR\\
\midrule
EWC ($\lambda_\textrm{EWC} = 1.0$)  & 33.0 & 38.1 & 17.6 & 30.3\\
EWC ($\lambda_\textrm{EWC} = 0.1$) & 32.9 & 38.2 & 17.1 & 30.6\\
EWC ($\lambda_\textrm{EWC} = 0.01$) & 32.2 & 37.7 & 15.9 & 29.8\\
EWC ($\lambda_\textrm{EWC} = 0.001$) & 31.8 & 36.7 & 17.3 & 29.7\\
\midrule
mEWC ($\lambda_\textrm{EWC} = 1.0$) & 33.0 & 38.5 & 16.5 & 30.5\\
mEWC ($\lambda_\textrm{EWC} = 0.1$) & 33.0 & 38.4 & 16.6 & 30.5\\
mEWC ($\lambda_\textrm{EWC} = 0.01$) & 32.2 & 36.6 & 19.1 & 29.8\\
mEWC ($\lambda_\textrm{EWC} = 0.001$) & 30.3 & 33.9 & 19.3 & 28.8\\
\bottomrule
\end{tabular}}
\end{center}
\vspace{-1em}
\caption{A comparison between the EWC with the full Fisher matrix VS the utilized mEWC with mean Fisher matrix on NIFF-DEFRCN on MS-COCO ($10$-shot). Further, a study on the impact of scaling EWC/mEWC regularization term.}
\label{tab:ewc}
\vspace{-1em}
\end{table}

In~\cref{tab:ewc}, we  show the difference between the vanilla EWC~\cite{ewc} with the full Fisher matrix and the utilized mean EWC (mEWC) using a mean Fisher matrix per parameter. Moreover, we show the effect of various scaling factors $\lambda_\textrm{EWC}$ when applying the EWC/mEWC penalty term during novel training. The lower the scaling factor, the more changes in parameters are allowed. Firstly, we observe that EWC can better maintain the base performance along different scaling factors at the price of the novel performance. Secondly, as we reduce the scaling factor for mEWC, the base performance drops compared to EWC. However, we achieve the same AP as EWC at $\lambda_\textrm{EWC}=0.01$ with a lower bAP and a higher nAP. Hence, we use this setting across our experiments. The user can decide on the bAP and nAP trade-off according to the application. Finally, we observe that in mEWC, the nAP improves with lower $\lambda_\textrm{EWC}$, however, at the cost of lower bAP.        
\subsection{Novel Training Loss Components}
\begin{table}[h!]
\begin{center}
\scalebox{0.9}{
\begin{tabular}{l|ccc|c}
	\toprule
  \multirow{2}{*}{Model Configuration} & \multicolumn{4}{c}{10-Shot Inference}\\
  &  AP & bAP & nAP & AR\\
\midrule
DeFRCN w/G.  & 30.7 & 35.0 & 17.7 & 28.8\\
\midrule
w/o $\mathcal{L}_\textrm{conf}$ & 28.9 & 32.6 & 17.7 & 26.4\\
 $\mathcal{L}_\textrm{conf}$ using KL & 30.7 & 35.0 & 17.7 & 28.2\\
\midrule
w/o Weighted feature terms & 29.8 & 33.9 & 17.6 & 28.0\\
w/o  L1 Reg. term & 30.5 & 34.8 & 17.8 & 28.6\\
\bottomrule
\end{tabular}}
\end{center}
\vspace{-1.5em}
\caption{Impact of various finetuning loss components on MS-COCO ($10$-shot). We finetune DeFRCN without base data using the generator without any regularization techniques.}
\label{tab:finetuning_losses}
\vspace{-1em}
\end{table}
In~\cref{tab:finetuning_losses}, we study the impact of various finetuning loss components. In the first row, we start with finetuning DeFRCN with the proposed generator and the novel training loss proposed in the main paper $\mathcal{L}_\textrm{N}$  without any regularization (i.e., CFA, data augmentations, and mEWC). Upon removing the cross-entropy confidence loss  $\mathcal{L}_\textrm{conf}$ (row2), we notice that the base performance drop by $2.4$ points causing the overall AP and the AR to drop. This denotes that the confidence loss helps generating base base features with higher probabilities at the final softmax layer. If we replace the cross-entropy loss with KL divergence (row3) between the teacher and student logits, we get the same results but with a slight drop in the AR. As for the distillation terms, we show that by removing the weighted feature distillation terms (row 4), the base and overall performance drops. Finally, if the proposed L1 regression distillation term is removed (row 5), we notice a slight decrease in the base  and overall performance. Motivated by this ablation, we opt to perform novel training with the overall loss containing the confidence and feature distillation loss terms.       


\section{Generator Training Analysis}
\begin{figure*}
     \begin{subfigure}[b]{0.32\textwidth}
         \includegraphics[width=\textwidth]{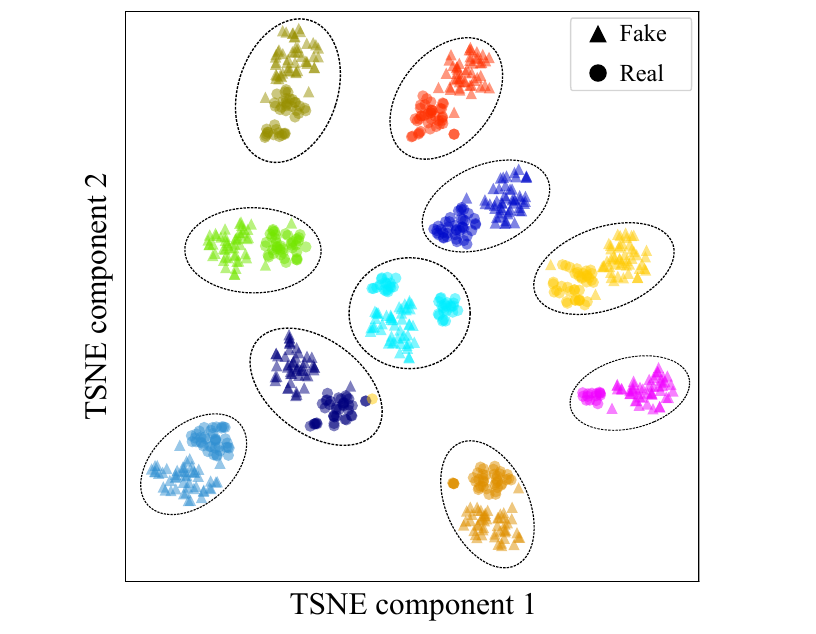}
         \caption{A TSNE visualization of the generated fake features via class-wise stats and real base features for $10$ random classes with $30$ features per class to better illustrate the forged features distribution.}
         \label{fig:gen_tsne}
     \end{subfigure}
     \hfill
     \begin{subfigure}[b]{0.32\textwidth}
         \hspace{1.0em}
         \includegraphics[height=0.75\textwidth]{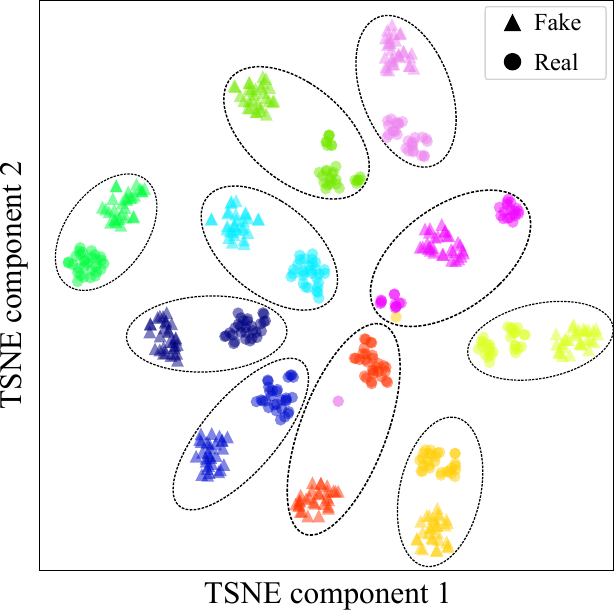}
         \caption{A TSNE visualization of the generated fake features via class-agnostic stats and real base features for $10$ random classes with $30$ features per class.}
         \label{fig:gen_tsne_agn}
     \end{subfigure}
     \hfill
     \begin{subfigure}[b]{0.32\textwidth}
         \includegraphics[width=\textwidth]{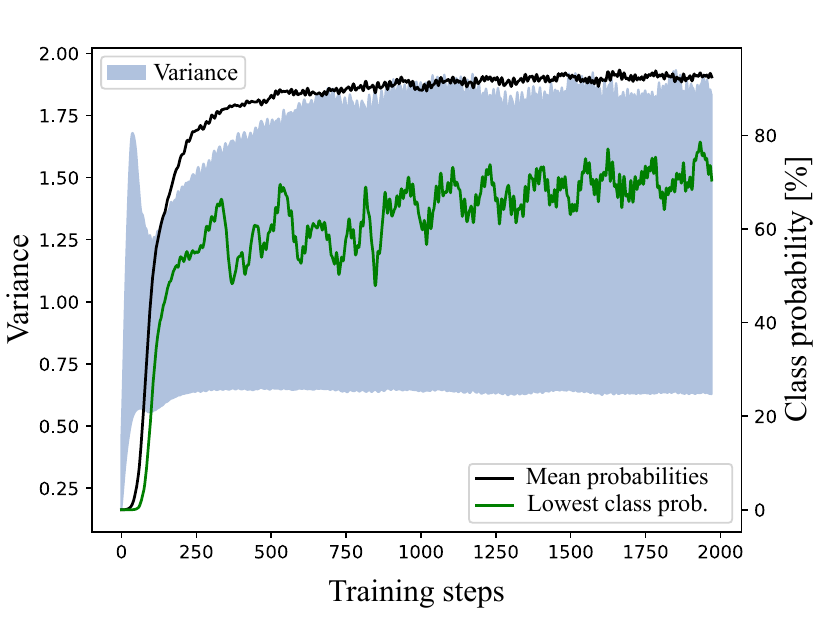}
         \caption{A highlight of the fake class probabilities along with the features variance. We show the lowest class probability and the mean probabilities across all base classes (MS-COCO).}
         \label{fig:gen_train_plot}
     \end{subfigure}
     \vspace{-1em}
        \caption{A TSNE visualization of some generated features via class-wise (a) and class-agnostic (b) stats. The fake class probabilities as well as the features' variance during generator training is illustrated in (c).}
        \label{fig:three graphs}
\end{figure*}


\begin{table*}[h!]
\scriptsize
    \begin{tabular}{c| c | c | c c c c c | c c c c c | c c c c c}
      \Xhline{1pt}
      \multirow{2}{*}{\textbf{Methods} / \textbf{Shots}} &
      \multirow{2}{*}{\textbf{w/E}} &
      \multirow{2}{*}{\textbf{w/B}} &
      \multicolumn{5}{c |}{\textbf{Novel Set 1}}  &
      \multicolumn{5}{c |}{\textbf{Novel Set 2}}  &
      \multicolumn{5}{c}{\textbf{Novel Set 3}}\\
      & & & 1 &  {2} & {3} & {5} &  {10} & 1 &  {2} & {3} & {5} &  {10} & 1 &  {2} & {3} & {5} &  {10} \\ \Xhline{1pt}
      FRCN-ft-full\cite{TFA} & \xmark & \cmark & 15.2 & 20.3 & 29.0 & 25.5 & 28.7 & 13.4 & 20.6 & 28.6 & 32.4 & 38.8 & 19.6 & 20.8 & 28.7 & 42.2 & 42.1 \\
      TFA w/ fc\cite{TFA}& \xmark & \cmark & 36.8 & 29.1 & 43.6 & 55.7 & 57.0 & 18.2 & 29.0 & 33.4 & 35.5 & 39.0 & 27.7 & 33.6 & 42.5 & 48.7 & 50.2 \\
      TFA w/ cos\cite{TFA}& \xmark & \cmark & 39.8 & 36.1 & 44.7 & 55.7 & 56.0 & 23.5 & 26.9 & 34.1 & 35.1 & 39.1 & 30.8 & 34.8 & 42.8 & 49.5 & 49.8 \\
      MPSR\cite{MPSR}& \xmark & \cmark & 42.8 & 43.6 & 48.4 & 55.3 & 61.2 & 29.8 & 28.1 & 41.6 & 43.2 & 47.0 & 35.9 & 40.0 & 43.7 & 48.9 & 51.3 \\
      DeFRCN\cite{defrcn}& \xmark & \cmark & 57.0 & 58.6 & 64.3 & 67.8 & 67.0 &
     35.8 & 42.7 & 51.0 & 54.4 & 52.9 & 52.5 & 56.6 & 55.8 & \second{60.7} & 62.5 \\ \Xhline{1pt}
      Meta R-CNN$^{*}$\cite{MetaRCNN}& \xmark & \cmark & 16.8 & 20.1 & 20.3 & 38.2 & 43.7 & 7.7 & 12.0 & 14.9 & 21.9 & 31.1 & 9.2 & 13.9 & 26.2 & 29.2 & 36.2 \\
      FSRW\cite{FSRW}& \xmark & \cmark & 14.8 & 15.5 & 26.7 & 33.9 & 47.2 & 15.7 & 15.3 & 22.7 & 30.1 & 39.2 & 19.2 & 21.7 & 25.7 & 40.6 & 41.3 \\
      MetaDet\cite{MetaDet}& \xmark & \cmark & 18.9 & 20.6 & 30.2 & 36.8 & 49.6 & 21.8 & 23.1 & 27.8 & 31.7 & 43.0 & 20.6 & 23.9 & 29.4 & 43.9 & 44.1\\
      FsDetView$^*$\cite{FsDetView}& \xmark & \cmark & 25.4 & 20.4 & 37.4 & 36.1 & 42.3 & 22.9 & 21.7 & 22.6 & 25.6 & 29.2 & 32.4 & 19.0 & 29.8 & 33.2 & 39.8 \\ \Xhline{1pt}
      CFA w/ fc~\cite{cfa}& \xmark & \cmark & 40.0 & 35.5 & 40.9 & 54.1 & 56.9 & 22.2 & 27.1 & 35.2 & 38.5 & 40.9 & 29.7 & 35.1 & 39.5 & 47.2 & 51.3 \\
      CFA w/ cos~\cite{cfa}& \xmark & \cmark & 41.2 & 43.6 & 49.5 & 56.5 & 57.3 & 21.3 & 27.4 & 35.3 & 39.1 & 42.1 & 31.7 & 39.1 & 44.6 & 49.9 & 52.6 \\
    CFA-DeFRCN~\cite{cfa}& \xmark & \cmark & \second{58.2} & \second{63.3} & \second{65.8} & \second{68.9} & \second{67.1} & \second{37.1} & \best{45.5} & \second{51.3} & \second{55.2} & \second{53.8} & \second{54.7} & \second{57.8} & \second{56.9} & 60.0 & \second{63.3} \\
     \Xhline{1pt}
      DeFRCN& \xmark & \xmark & 53.3 & 47.4 & 58.7 & 58.8 & 59.6 & 33.0 & 37.0 & 49.5 & 53.8 & 48.5 & 47.1 & 45.8 & 52.7 & 52.8 & 52.6 \\
      \rowcolor[HTML]{EFEFEF}
    NIFF-DeFRCN& \xmark & \xmark & \best{62.8} & \best{67.2} & \best{68.0} & \best{70.3} & \best{68.8} & \best{38.4} & \second{42.9} & \best{54.0} & \best{56.4} & \best{54.0} & \best{56.4} & \best{62.1} & \best{61.2} & \best{64.1} & \best{63.9} \\
      \midrule[1.5pt]
      Retentive R-CNN\cite{retentive}& \cmark & \cmark & 42.4 & 45.8 & 45.9 & 53.7 & 56.1 & 21.7 & 27.8 & 35.2 & 37.0 & 40.3 & 30.2 & 37.6 & 43.0 & 49.7 & 50.1 \\
      \Xhline{1pt}
      CFA w/ fc~\cite{cfa}& \cmark & \cmark &39.0 & 34.9 & 41.4 & 54.8 & 57.0 & 21.8 & 26.1 & 35.3 & 37.1 & 40.1 & 29.9 & 34.3 & 40.1 & 47.0 & 52.6 \\
      CFA w/ cos~\cite{cfa}& \cmark & \cmark &  42.4 &  43.9 & 50.3 & 56.6 & 57.3 & 21.0 & 27.5 & 35.3 & 38.6 & 41.4 & 32.3 &  38.0 & 44.5 & 49.8 & 52.7 \\
      CFA-DeFRCN~\cite{cfa}& \cmark & \cmark & \second{59.0} & \second{63.5} & \second{66.4} & \second{68.4} & \second{68.3} & \second{37.0} & \best{45.8} & \second{50.0} & \second{54.2} & \second{52.5} & \second{54.8} & \second{58.5} & \second{56.5} & \second{61.3} & \second{63.5} \\
      \Xhline{1pt}
      \rowcolor[HTML]{EFEFEF}
    NIFF-DeFRCN& \cmark & \xmark & \best{63.5} & \best{67.2} & \best{68.3} & \best{71.1} & \best{69.3} & \best{37.8} & \second{41.9} & \best{53.4} & \best{56.0} & \best{53.5} & \best{55.3} & \best{60.5} & \best{61.1} & \best{63.7} & \best{63.9} \\
      \Xhline{1pt}
    \end{tabular}
    \caption{PASCAL-VOC G-FSOD (nAP50) results for $1,2,3,5,10$-shot settings for all three splits are reported. Similar to~\cite{retentive,cfa}, w/E denotes the ensemble-based inference paradigm. The {\color{red}best} and {\color{blue}second-best} results are color coded. w/B indicates whether the base data is available during novel finetuning. '-' represents unrecorded results in prior works. '*' represents results reported in \cite{cfa}.}
    \vspace{-2em}
    \label{tab:voc-novel}
\end{table*}
In~\cref{fig:gen_tsne}, we show a TSNE visualization of the real and fake generated instance-level features for $10$ random MS-COCO base classes. We generate $30$ features per class to better visualize the generated feature distribution. We show that the forged features are consistently near the real base features (with some overlaps) confirming that the generator is able to depict near-real base feature distribution. Furthermore, we show the features generated using class-agnostic statistics in~\cref{fig:gen_tsne_agn}. In contrast to the features generated via the class-wise stats, the fake samples are farther away from the real ones.

In~\cref{fig:gen_train_plot}, we highlight the quality and diversity of the generated features with respect to the feature variance. We show the mean class probability (black) across all classes for the generated features and the lowest class probability (green). It can be seen that the generator can learn diverse features with a high variance while maintaining high class probabilities with $95\%$ mean class probability and around $75\%$ for the lowest class probability. 

\section{PASCAL-VOC Novel Results}

We report the novel performance on PASCAL-VOC (nAP50) in~\cref{tab:voc-novel}. We show that adopting NIFF achieves state-of-the-art results with and without the ensemble evaluation protocol. As previously mentioned, although that the performance on novel classes is not our primary objective, NIFF-DeFRCN achieves the state-of-the-art on PASCAL-VOC in all cases but the $2$-shot setting in split $2$ .
\section{Multiple Runs}
\begin{table*}[h!]
    \centering
    \scalebox{0.9}{
    \begin{tabular}{c | c c c | c c c | c c c}
      \Xhline{1pt}
      \multirow{2}{*}{\textbf{Methods} / \textbf{Shots}} & 
      \multicolumn{3}{c |}{\textbf{5 shot}}  &
      \multicolumn{3}{c |}{\textbf{10 shot}}  &
      \multicolumn{3}{c}{\textbf{30 shot}}\\
      & \textbf{AP} &  \textbf{bAP} & \textbf{nAP} & \textbf{AP} &  \textbf{bAP} & \textbf{nAP} & \textbf{AP} &  \textbf{bAP} & \textbf{nAP} \\ \Xhline{1pt}
      TFA w/ fc\cite{TFA} & 25.6$\pm$0.5 & 31.8$\pm$0.5 & 6.9$\pm$0.7 & 26.2$\pm$0.5 & 32.0$\pm$0.5 & 9.1$\pm$0.5 & 28.4$\pm$0.3 & 33.8$\pm$0.3 & 12.0$\pm$0.4 \\
      
      TFA w/ cos\cite{TFA} & 25.9$\pm$0.6 & 32.3$\pm$0.6 & 7.0$\pm$0.7 & 26.6$\pm$0.5 & 32.4$\pm$0.6 & 9.1$\pm$0.5 & 28.7$\pm$0.4 & 34.2$\pm$0.4 & 12.1$\pm$0.4 \\
      
      CFA w/ fc~\cite{cfa}  & 29.1$\pm$0.3 & 36.2$\pm$0.3 & 7.7$\pm$0.6 & 29.9$\pm$0.3 & 36.7$\pm$0.2& 9.6$\pm$0.6 & 30.8$\pm$0.2 & 36.6$\pm$0.2 & 13.6$\pm$0.3\\
      
      CFA w/ cos~\cite{cfa}  & 29.3$\pm$0.2 & 36.0$\pm$0.2 & 9.2$\pm$0.5& 30.2$\pm$0.2 & 36.6$\pm$0.1& 11.2$\pm$0.5 & 31.1$\pm$0.1 & 36.6$\pm$0.1 & 14.8$\pm$0.2\\
      
      DeFRCN\cite{defrcn} & 27.8$\pm$0.3 & 32.6$\pm$0.3 & 13.6$\pm$0.7 & 29.7$\pm$0.2 & 34.0$\pm$0.2 & 16.8$\pm$0.6 & 31.4$\pm$0.1 & 34.8$\pm$0.1 & 21.2$\pm$0.4\\
        
      CFA-DeFRCN~\cite{cfa} & 28.4$\pm$0.2 & 32.8$\pm$0.2 & 15.2$\pm$0.5 & 30.2$\pm$0.2 & 34.0$\pm$0.2 & 18.8$\pm$0.4 & 31.7$\pm$0.1 & 34.6$\pm$0.1 & 23.0$\pm$0.3\\

      \rowcolor[HTML]{EFEFEF}
      \cellcolor{white} NIFF-DeFRCN & 31.1$\pm$0.1 & 36.6$\pm$0.0 & 14.6$\pm$0.2 & 32.1$\pm$0.1 & 36.8$\pm$0.1 & 18.0$\pm$0.2 & 33.3$\pm$0.0 & 37.7$\pm$0.1 & 20.0$\pm$0.1\\
      \hline
    \end{tabular}}
    \caption{G-FSOD experimental results for 5,10,30-shot settings on MS-COCO. We report base (bAP), novel (nAP), and overall (AP) for multiple runs using $10$ different seeds. }
    \label{tab:coco_mult_runs}
\end{table*}
\begin{table*}[h!]
\small
    \centering
    \begin{tabular}{c|c| c c c c c }
      \Xhline{1pt}
      \multirow{2}{*}{\textbf{Set}} &
      \multirow{2}{*}{\textbf{Methods}} &
      \multicolumn{5}{c}{\textbf{Shots}} \\
       & &1&2&3&5& 10 \\ \Xhline{1pt}
      & CFA w/ fc~\cite{cfa}& 66.3$\pm$0.8 & 68.0$\pm$0.5 & 70.1$\pm$0.4 & 71.7$\pm$0.5 & 73.2$\pm$0.5 \\
      & CFA w/ cos~\cite{cfa}& 66.5$\pm$0.9 & 69.2$\pm$0.6 & 71.1$\pm$0.6 & 72.5$\pm$0.4 & 73.4$\pm$0.4 \\
      & DeFRCN~\cite{defrcn} & 67.8$\pm$1.4 & 71.3$\pm$0.8 & 72.6$\pm$0.5 & 73.6$\pm$0.5 & 74.1$\pm$0.5 \\
      & CFA-DeFRCN~\cite{cfa} & 69.0$\pm$1.4 & 72.6$\pm$0.7 & 73.1$\pm$0.4 & 74.0$\pm$0.5 & 74.3$\pm$0.4 \\
      \rowcolor[HTML]{EFEFEF}
        \cellcolor{white} \multirow{-5}{*}{\textbf{All Set 1}} &  NIFF-DeFRCN & 71.2$\pm$0.8 & 74.2$\pm$0.4 & 75.4$\pm$0.4 & 76.3$\pm$0.3 & 76.7$\pm$0.3 \\
       \hline
       & CFA w/ fc~\cite{cfa} & 64.9$\pm$0.9 & 66.4$\pm$0.7 & 68.3$\pm$0.5 & 69.6$\pm$0.3 & 70.8$\pm$0.5 \\
       & CFA w/ cos~\cite{cfa} & 64.1$\pm$0.9 & 66.5$\pm$0.5 & 68.1$\pm$0.5 & 69.3$\pm$0.2 & 70.4$\pm$0.4 \\
       & DeFRCN~\cite{defrcn} & 65.2$\pm$1.0 & 68.0$\pm$0.8 & 69.2$\pm$0.6 & 70.6$\pm$0.6 & 71.3$\pm$0.5 \\
       & CFA-DeFRCN~\cite{cfa} & 66.4$\pm$1.0 & 69.0$\pm$0.8 & 70.4$\pm$0.7 & 71.3$\pm$0.7 & 72.1$\pm$0.4 \\ 
       \rowcolor[HTML]{EFEFEF}
       \cellcolor{white} \multirow{-5}{*}{\textbf{All Set 2}} & NIFF-DeFRCN & 68.0$\pm$0.8 & 70.5$\pm$0.5 & 71.7$\pm$0.5 & 72.8$\pm$0.4 & 73.7$\pm$0.3 \\ 
       \hline
       & CFA w/ fc~\cite{cfa} & 65.2$\pm$0.8 & 66.8$\pm$0.8 & 69.1$\pm$0.7 & 70.9$\pm$0.6 & 72.3$\pm$0.4 \\
       & CFA w/ cos~\cite{cfa} & 64.9$\pm$1.2 & 67.5$\pm$1.0 & 69.7$\pm$0.8 & 71.6$\pm$0.5 & 72.7$\pm$0.3 \\
       & DeFRCN~\cite{defrcn} & 66.9$\pm$2.0 & 70.6$\pm$0.8 & 71.2$\pm$0.6 & 72.9$\pm$0.5 & 73.5$\pm$0.3 \\
       & CFA-DeFRCN~\cite{cfa} & 68.3$\pm$1.6 & 71.4$\pm$0.8 & 72.3$\pm$0.5 & 73.5$\pm$0.5 & 74.0$\pm$0.3 \\
       \rowcolor[HTML]{EFEFEF}
       \cellcolor{white}\multirow{-5}{*}{\textbf{All Set 3}} & NIFF-DeFRCN & 70.7$\pm$0.7 & 73.7$\pm$0.5 & 74.7$\pm$0.4 & 75.5$\pm$0.3 & 76.3$\pm$0.2 \\
       \hline
    \end{tabular}
    \caption{G-FSOD experimental results for 1,2,3,5,10-shot settings on the three all sets of Pascal VOC (AP50).}
    \label{tab:voc_all_mult_runs}
\end{table*}
\begin{table*}[h!]
\small
    \centering
    \begin{tabular}{c| c | c c c c c }
      \Xhline{1pt}
      \multirow{2}{*}{\textbf{Set}} &
      \multirow{2}{*}{\textbf{Methods}} &
      \multicolumn{5}{c}{\textbf{Shots}} \\
       & & 1 &  {2} & {3} & {5} &  {10} \\ \Xhline{1pt}
      & CFA w/ fc~\cite{cfa}& 28.2$\pm$3.1 & 35.0$\pm$1.9 & 41.9$\pm$1.4 & 47.8$\pm$1.6 & 53.3$\pm$1.6 \\
      & CFA w/ cos~\cite{cfa}& 30.9$\pm$3.9 & 40.9$\pm$2.5 & 47.8$\pm$2.4 & 53.1$\pm$1.4 & 56.1$\pm$1.4 \\
      & DeFRCN~\cite{defrcn} & 43.8$\pm$4.3 & 57.5$\pm$2.5 & 61.4$\pm$1.7 & 65.3$\pm$0.9 & 67.0$\pm$1.4 \\
      & CFA-DeFRCN~\cite{cfa} & 45.4$\pm$4.9 & 60.3$\pm$2.2 & 62.1$\pm$1.4 & 66.4$\pm$0.9 & 67.6$\pm$1.2\\
      \rowcolor[HTML]{EFEFEF}
      \cellcolor{white}\multirow{-5}{*}{\textbf{Novel Set 1}} & NIFF-DeFRCN & 46.0$\pm$3.0 & 57.2$\pm$1.7 & 62.0$\pm$1.4 & 65.5$\pm$1.1 & 67.2$\pm$1.1 \\
       \hline
       & CFA w/ fc~\cite{cfa} & 20.0$\pm$3.5 & 26.4$\pm$2.9 & 32.8$\pm$2.2 & 37.3$\pm$1.7 & 41.8$\pm$1.9 \\
       & CFA w/ cos~\cite{cfa} & 21.0$\pm$3.5 & 29.0$\pm$2.3 & 34.6$\pm$2.3 & 38.9$\pm$1.2 & 43.0$\pm$1.9 \\
       & DeFRCN\cite{defrcn} & 31.5$\pm$3.6 & 40.9$\pm$2.2 & 45.6$\pm$2.0 & 50.1$\pm$1.4 & 52.9$\pm$1.1 \\ 
       & CFA-DeFRCN~\cite{cfa} & 32.9$\pm$3.7 & 42.3$\pm$2.2 & 47.1$\pm$1.9 & 51.2$\pm$1.4 & 55.3$\pm$1.3 \\ 
       \rowcolor[HTML]{EFEFEF}
       \cellcolor{white}\multirow{-5}{*}{\textbf{Novel Set 2}} & NIFF-DeFRCN & 30.1$\pm$3.0 & 39.6$\pm$1.8 & 45.0$\pm$1.9 & 49.4$\pm$1.6 & 52.8$\pm$1.3 \\ 
       \hline
       & CFA w/ fc~\cite{cfa} & 20.3$\pm$3.4 & 26.4$\pm$3.1 & 34.3$\pm$2.5 & 41.2$\pm$2.4 & 46.5$\pm$1.6 \\
       & CFA w/ cos~\cite{cfa} & 21.5$\pm$4.7 & 30.4$\pm$4.1 & 38.4$\pm$2.8 & 45.5$\pm$2.1 & 49.9$\pm$1.0 \\
       & DeFRCN~\cite{defrcn} & 38.2$\pm$6.8 & 50.9$\pm$2.8 & 54.1$\pm$2.2 & 59.2$\pm$1.2 & 61.9$\pm$1.3 \\
       & CFA-DeFRCN~\cite{cfa} & 41.4$\pm$5.8 & 52.9$\pm$3.0 & 56.1$\pm$1.7 & 60.3$\pm$1.1 & 62.9$\pm$0.9 \\
       \rowcolor[HTML]{EFEFEF}
       \cellcolor{white}\multirow{-5}{*}{\textbf{Novel Set 3}} & NIFF-DeFRCN & 41.1$\pm$2.6 & 52.5$\pm$1.8 & 56.4$\pm$1.5 & 59.7$\pm$1.2 & 62.1$\pm$1.0 \\
       \hline
    \end{tabular}
    \vspace{-0.5em}
    \caption{G-FSOD experimental results for 1,2,3,5,10-shot settings on the three novel sets of Pascal VOC (nAP50).}
    \label{tab:voc_novel_mult_runs}
    \vspace{-1.5em}
\end{table*}
We run NIFF-DeFRCN using $10$ and $30$ different seeds for MS-COCO and PASCAL-VOC, respectively. Then, we compare with the baselines~\cite{TFA,defrcn,cfa}. The aim is to investigate the performance robustness over multiple runs.

\textbf{MS-COCO.} In~\cref{tab:coco_mult_runs}, we show the results for MS-COCO dataset. We notice that despite the absence of base data, NIFF-DeFRCN consistently achieves a higher AP and bAP over all shot settings.

\textbf{PASCAL-VOC.} In~\cref{tab:voc_all_mult_runs} and~\cref{tab:voc_novel_mult_runs}, we demonstrate the AP50 and nAP50 results, respectively, for PASCAL-VOC dataset. Similar to MS-COCO, we observe that NIFF-DeFRCN consistently achieves a higher AP50 while delivering competitive results on the nAP50 over various shots.

\addtolength{\tabcolsep}{-4.5pt}    

\setlength{\mywidth}{0.2\textwidth}

\bgroup
\def\arraystretch{0.5}
\begin{figure*}[t!]
\begin{tabular} {ccccc}

\includegraphics[width=\mywidth, height=\mywidth]{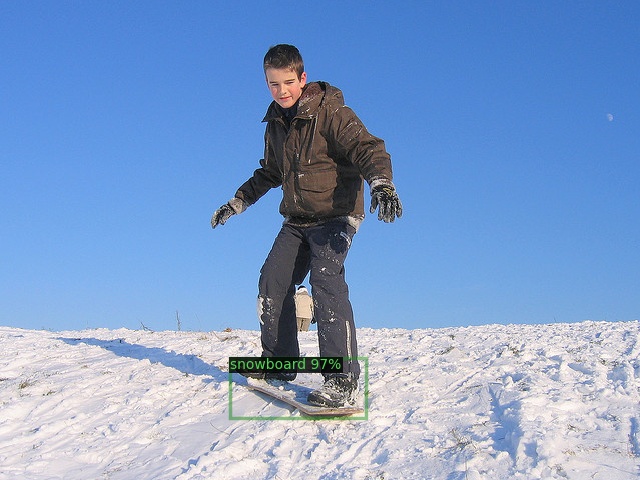} &  
\includegraphics[width=\mywidth, height=\mywidth]{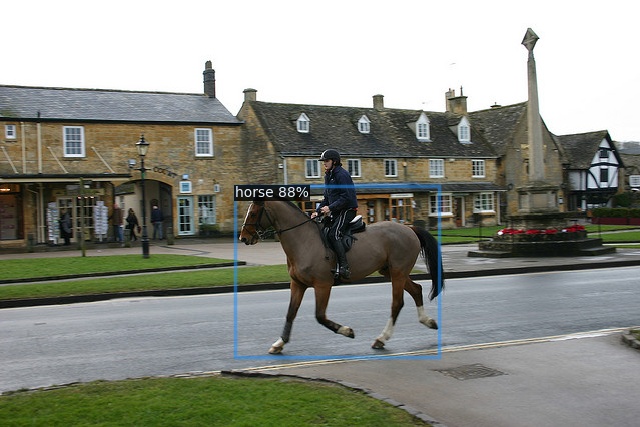} &   
\includegraphics[width=\mywidth, height=\mywidth]{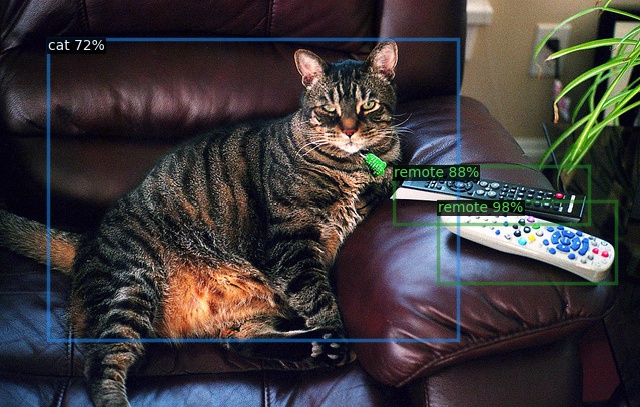} &   
\includegraphics[width=\mywidth, height=\mywidth]{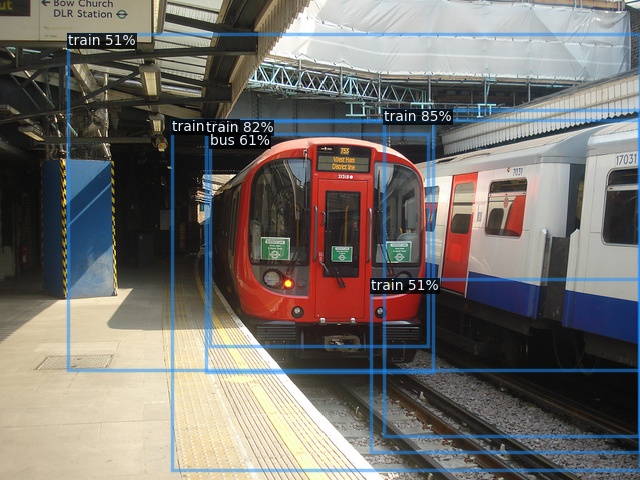} &
\includegraphics[width=\mywidth, height=\mywidth]{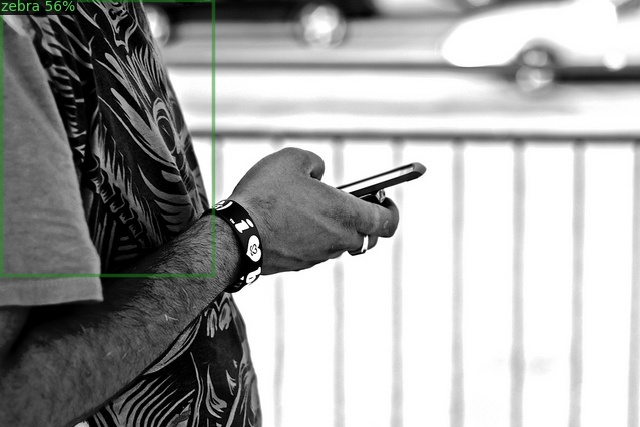} \\

\includegraphics[width=\mywidth, height=\mywidth]{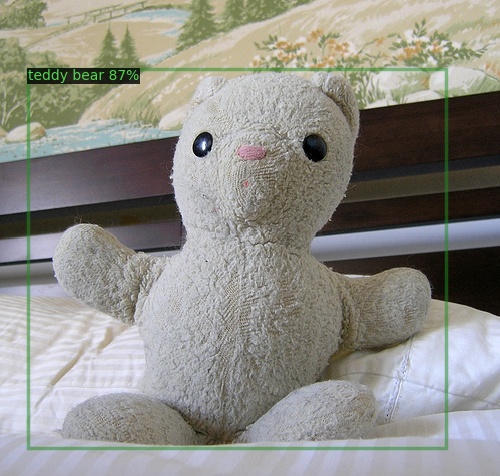} &  
\includegraphics[width=\mywidth, height=\mywidth]{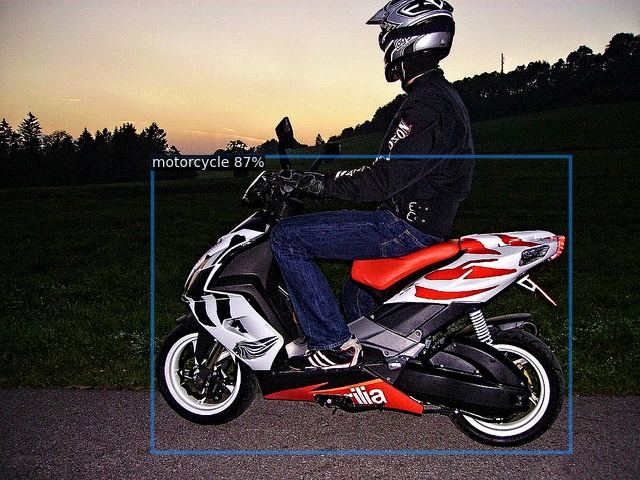} &   
\includegraphics[width=\mywidth, height=\mywidth]{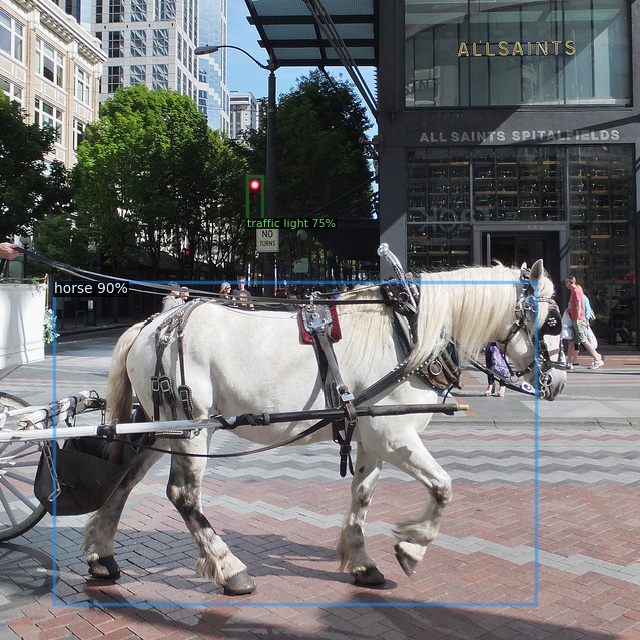} &  
\includegraphics[width=\mywidth, height=\mywidth]{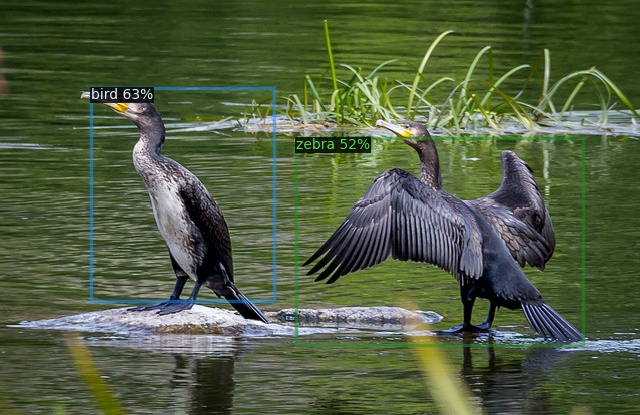} &   
\includegraphics[width=\mywidth, height=\mywidth]{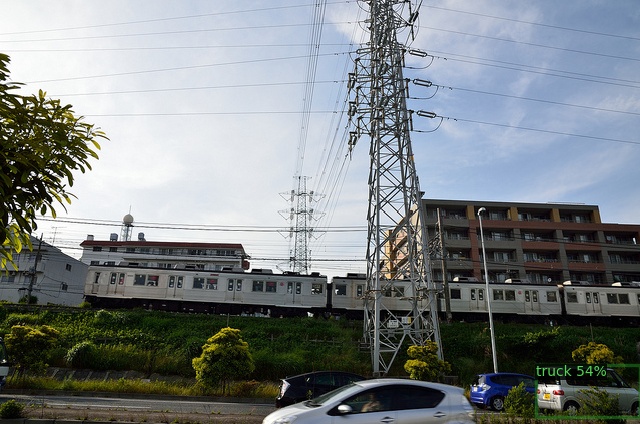} \\

\includegraphics[width=\mywidth, height=\mywidth]{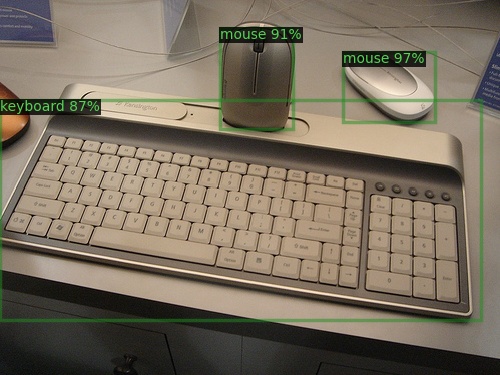} &  
\includegraphics[width=\mywidth, height=\mywidth]{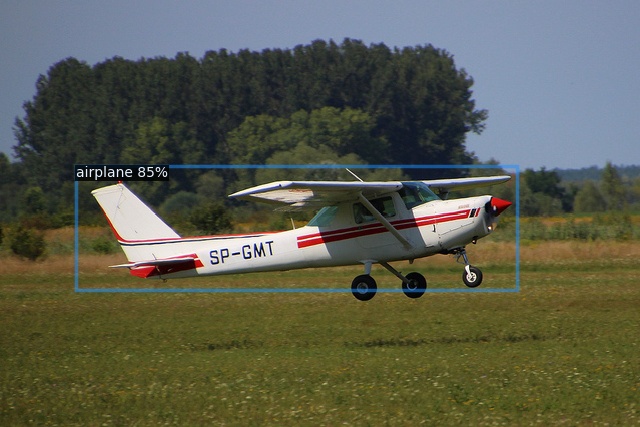} &   
\includegraphics[width=\mywidth, height=\mywidth]{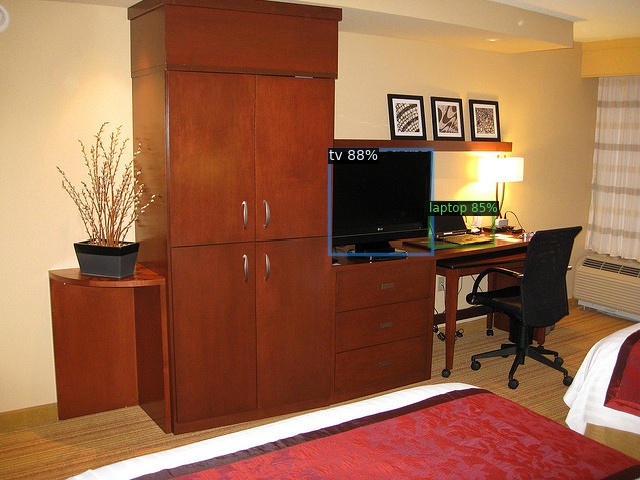} &  
\includegraphics[width=\mywidth, height=\mywidth]{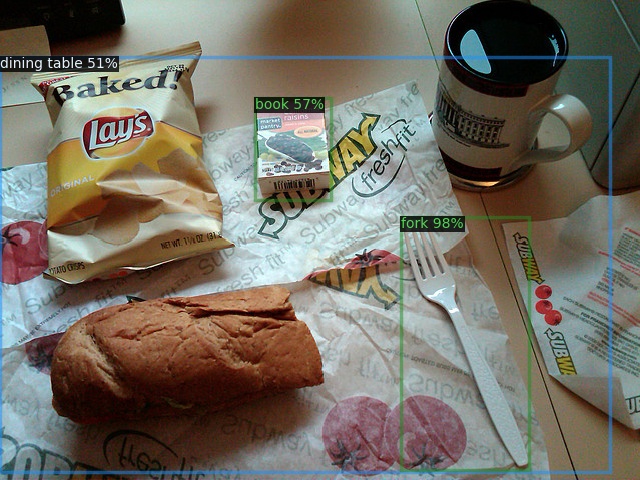} &   
\includegraphics[width=\mywidth, height=\mywidth]{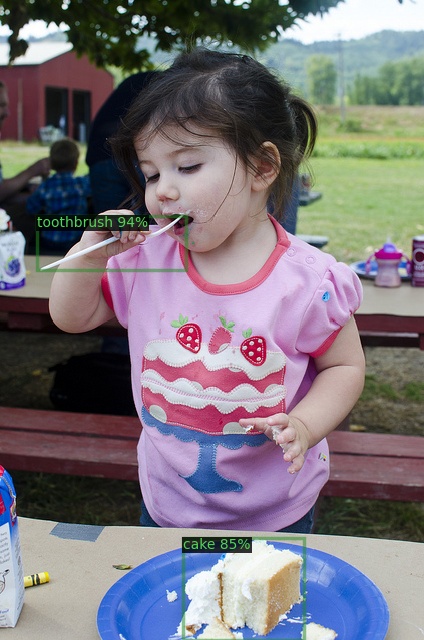} \\

\includegraphics[width=\mywidth, height=\mywidth]{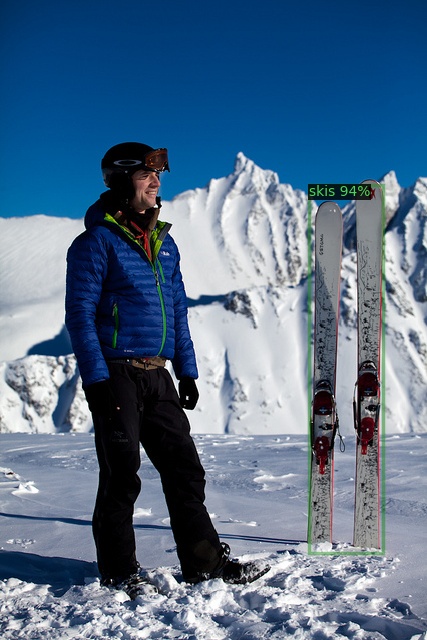} &  
\includegraphics[width=\mywidth, height=\mywidth]{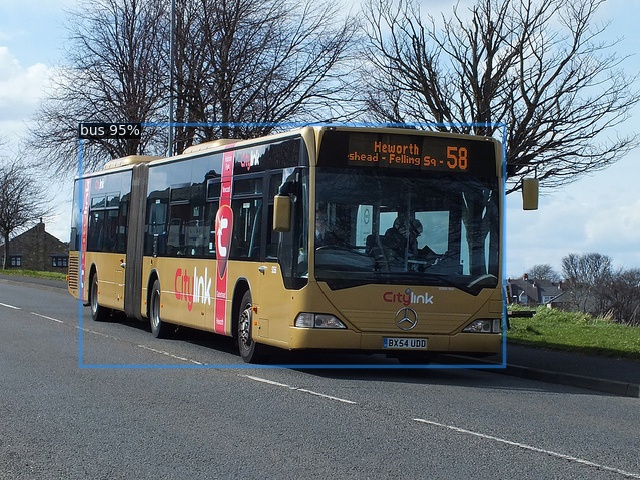} &
\includegraphics[width=\mywidth, height=\mywidth]{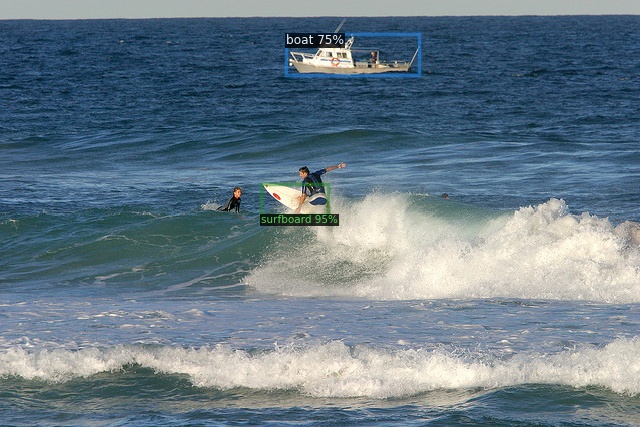} &   
\includegraphics[width=\mywidth, height=\mywidth]{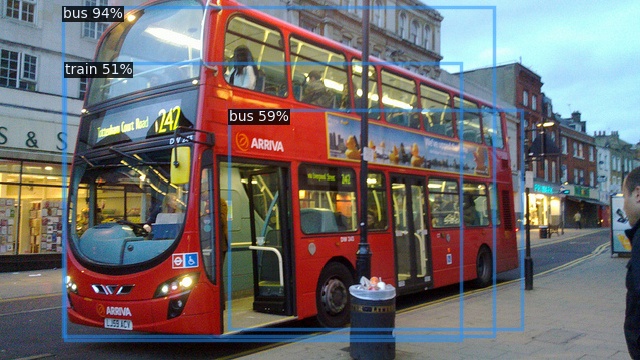} &  
\includegraphics[width=\mywidth, height=\mywidth]{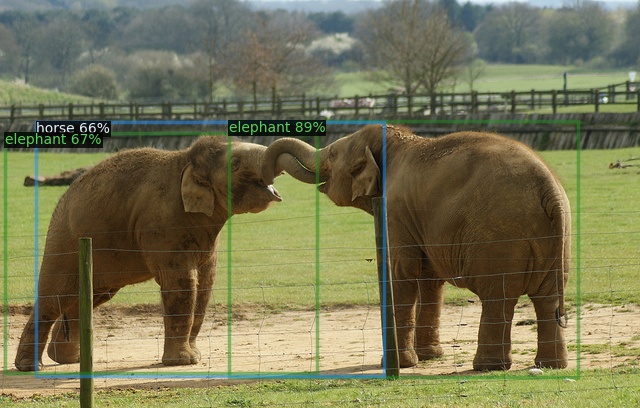} \\

\includegraphics[width=\mywidth, height=\mywidth]{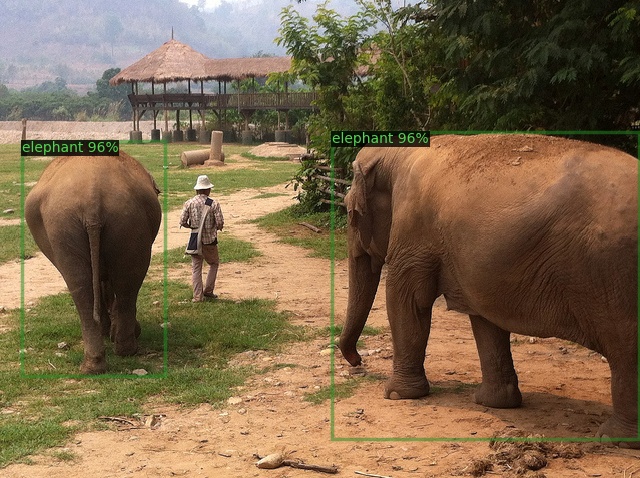} &  
\includegraphics[width=\mywidth, height=\mywidth]{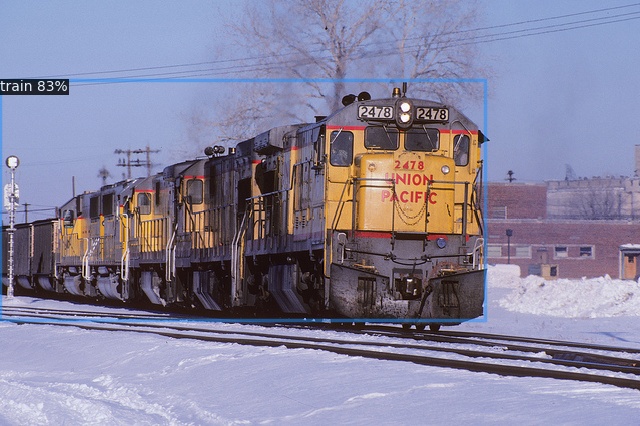} &
\includegraphics[width=\mywidth, height=\mywidth]{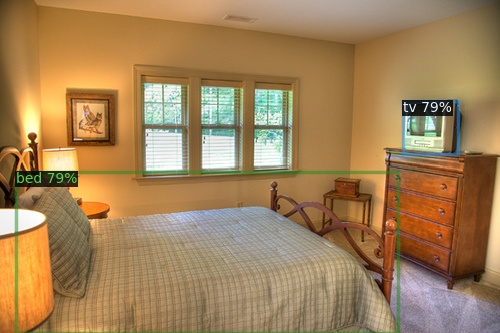} &   
\includegraphics[width=\mywidth, height=\mywidth]{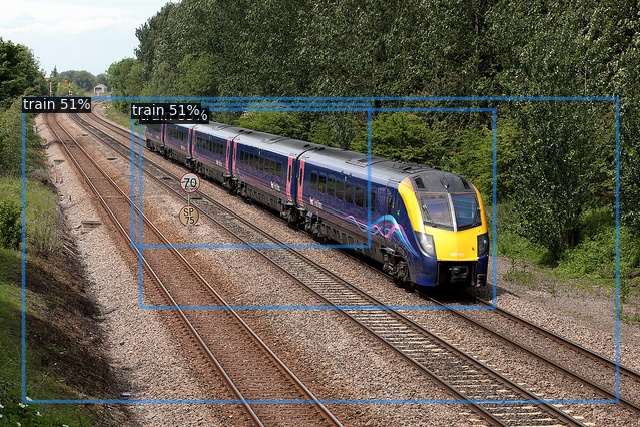} &  
\includegraphics[width=\mywidth, height=\mywidth]{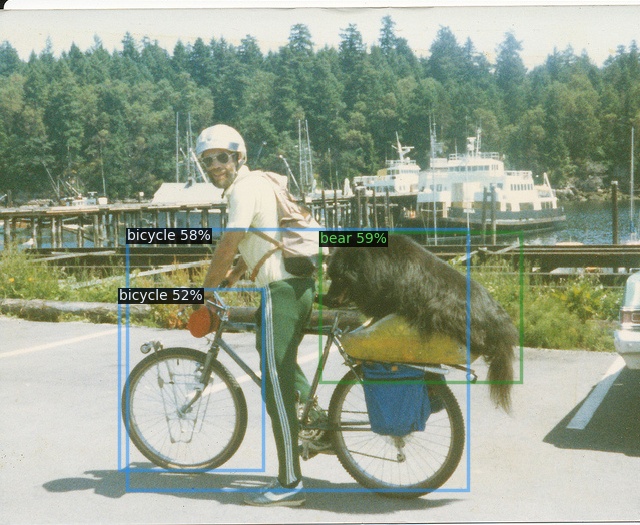} \\
 
\end{tabular}
	\caption{Qualitative results of the proposed NIFF method (NIFF-DeFRCN) on the MS-COCO($10$-shot) dataset. Success scenarios are demonstrated in the first three columns show while the last two columns present the failure scenarios. Base classes are denoted by green bounding boxes while novel classes are colored with blue. \vspace{-1em}}
	\label{fig::qualitative}
 \end{figure*}
 \egroup
 \addtolength{\tabcolsep}{4.5pt}
\section{Qualitative Results}

We present various qualitative results in~\cref{fig::qualitative} on the MS-COCO ($10$-shot). In the first column, we show images with only base classes (green boxes) followed by images with only novel classes (blue boxes) in the second column. In the third column, we present images with both classes. We chose to present these three cases to validate the performance of the proposed NIFF in various scenarios. . Further, we also present various failure cases in the last two columns. 
\clearpage
\clearpage
{\small
\bibliographystyle{ieee_fullname}
\bibliography{supp}
}